\documentclass[runningheads]{llncs}

\usepackage{eccv}

\usepackage{eccvabbrv}

\usepackage{booktabs}
\usepackage{multirow}
\usepackage{makecell}
\usepackage{graphicx}
\usepackage[table]{xcolor}
\usepackage{algorithm}
\usepackage{algpseudocode}
\usepackage{amssymb}
\usepackage{marvosym}
\usepackage{titletoc}

\usepackage[accsupp]{axessibility}  % Improves PDF readability for those with disabilities.

\usepackage{hyperref}
\hypersetup{hypertexnames=false}

\usepackage{orcidlink}

\newcommand{\myparagraph}[1]{\vspace{1pt}\noindent{\bf{#1}}~~}

\newcommand{\pmstd}[2]{%
  #1\kern0.05em{\raisebox{-0.00ex}{\tiny$\pm$#2}}%
}
\newcommand{\equalcontribmark}{\textsuperscript{*}}
\makeatletter
\def\@fnsymbol#1{\ensuremath{\ifcase#1\or\mbox{*}\or\mbox{\Letter}\or
   \star\or{\star\star}\or{\star\star\star}\or \dagger\or \ddagger\or
   \mathchar "278\or \mathchar "27B\or \|\or **\or \dagger\dagger
   \or \ddagger\ddagger \else\@ctrerr\fi}}
\makeatother

\begin{document}

% \title{Author Guidelines for ECCV Submission} 
\title{\textit{Together, Then Apart}: Balancing Alignment and Distinctiveness for Multimodal Survival Analysis}

\titlerunning{\textit{Together, Then Apart}}

\author{Wenjing Liu\inst{1,2}\orcidlink{0009-0004-0996-6773}\thanks{Equal contribution.} \and
Qin Ren\inst{1}\orcidlink{0009-0003-0156-3868}\equalcontribmark \and
Wen Zhang\inst{1,3}\orcidlink{0009-0004-2289-3943} \and \\
Yuewei Lin\inst{4}\orcidlink{0000-0002-1429-4543} \and
Chenyu You\inst{1}\orcidlink{0000-0001-8365-7822}\thanks{Corresponding author.}}

\authorrunning{W.~Liu et al.}

\institute{Stony Brook University, Stony Brook, NY, USA \and
Stanford University, Stanford, CA, USA \and
Johns Hopkins University, Baltimore, MD, USA \and
Brookhaven National Laboratory, Upton, NY, USA\\
\email{chenyu.you@stonybrook.edu}\\
\small{\url{https://y-research-sbu.github.io/TTA}}}

\maketitle

\begin{abstract}
Multimodal survival analysis aims to improve cancer prognosis using heterogeneous biomedical data, such as histopathology images and genomic profiles. A common strategy is to align representations across modalities so that shared signals can be captured. However, strong cross-modal alignment can also remove modality-specific evidence that is critical for survival prediction. In this paper, we revisit multimodal survival learning from a simple observation: effective models should first discover shared patterns across modalities, and then preserve modality-specific signals. This motivates a representation learning principle that we refer to as \textit{Together Then Apart}.
Based on this idea, we propose \textbf{TTA}, a framework that balances cross-modal alignment and representation distinctiveness. TTA first performs prototype-based alignment to capture shared survival-related structures between modalities. It then encourages modality-specific distinctiveness through an anchor-guided contrastive objective. To further account for modality imbalance and noisy correspondences, we model cross-modal interactions using unbalanced optimal transport.
We evaluate the proposed approach on multiple TCGA cancer cohorts with paired histopathology and genomic data. TTA consistently improves survival prediction over recent multimodal survival models. Moreover, the learned prototype structures reveal interpretable cross-modal patterns associated with clinical outcomes.
Code is available at \url{https://github.com/Y-Research-SBU/TTA}.
% Integrating heterogeneous modalities such as histopathology and genomics is central to advancing survival analysis, yet most existing methods prioritize cross-modal alignment through attention-based fusion mechanisms, often at the expense of modality-specific characteristics. This overemphasis on alignment leads to representation collapse and reduced diversity. In this work, we revisit multi-modal survival analysis via the dual lens of \textbf{alignment} and \textbf{distinctiveness}, positing that preserving modality-specific structure is as vital as achieving semantic coherence.
% Extensive experiments on five TCGA benchmarks show that TTA achieves state-of-the-art overall performance among multimodal survival methods. Beyond empirical gains, our formulation provides a new theoretical perspective of how alignment and distinctiveness can be jointly achieved in a robust, interpretable, and biologically meaningful multi-modal survival analysis. 
\end{abstract}

\section{Introduction}
\label{sec:intro}

\begin{figure}
\centering
\includegraphics[width=0.9\linewidth]{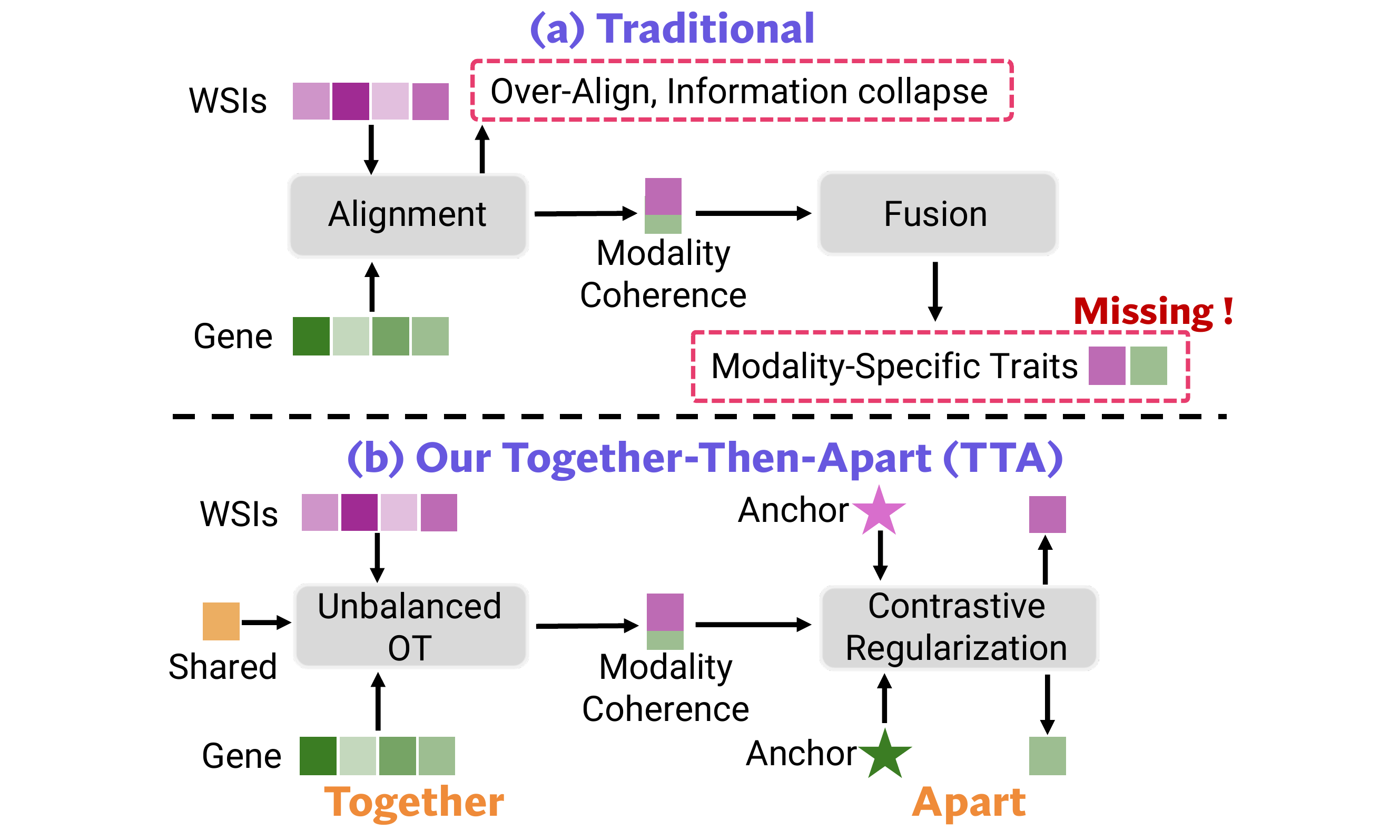}
\caption{\textbf{Traditional multimodal learning vs. our TTA framework.}
Conventional approaches align heterogeneous modalities in a shared representation space, which can suppress modality-specific signals. 
TTA follows a \textit{Together–Then–Apart} strategy: modalities are first aligned to capture shared patterns (\textit{Together}), and then separated to preserve modality-specific information (\textit{Apart}).}
\label{fig:intro}
\end{figure}

Multimodal survival analysis seeks to improve cancer prognosis by combining heterogeneous biomedical data sources. Among them, pathology whole-slide images (WSIs) and genomic profiles provide two complementary views of tumor biology. 
WSIs capture rich morphological patterns and histopathological biomarkers and are typically modeled using Multiple Instance Learning (MIL) frameworks that aggregate patch-level features into slide-level representations~\cite{ilse2018attention,shao2021transmil,yao2020whole,jackson2020single}. 
Genomic measurements, often derived from transcriptomic or pathway-level analyses~\cite{liberzon2015molecular,elmarakeby2021biologically,reimand2019pathway}, reveal molecular programs and mutation-driven signals that are not observable in tissue morphology~\cite{nunes2024prognostic}. 
Combining these modalities has therefore become a central direction in multimodal survival modeling~\cite{mobadersany2018predicting,chen2020pathomic,chen2021multimodal,chen2022pan,ding2023pathology,jaume2024modeling,xu2023multimodal}. 
Most recent methods adopt attention-based fusion mechanisms to learn joint representations that capture cross-modal correlations~\cite{chen2020pathomic,chen2021multimodal,jaume2024modeling,song2024multimodal}.

% Integrating multi-modal information, particularly pathology whole slide images (WSIs) and genomic profiles, has become increasingly central to survival analysis~\cite{acosta2022multimodal,chen2021multimodal,xu2023multimodal,jenkins2005survival,salerno2023high}. In the context of survival analysis, WSIs, typically partitioned into image patches and modeled through Multiple Instance Learning (MIL) frameworks~\cite{ilse2018attention,shao2021transmil,yao2020whole}, capture rich morphological patterns and histopathological biomarkers~\cite{jackson2020single}.
% In parallel, genomic data, often  processed via transcriptomics-based methods~\cite{liberzon2015molecular,elmarakeby2021biologically,reimand2019pathway}, reveal molecular signatures and mutation profiles essential to tumor characterization~\cite{nunes2024prognostic}.
% Together, these \textit{morphological} and \textit{molecular} modalities offer complementary prognostic perspectives, motivating a growing body of multi-modal survival analysis methods~\cite{mobadersany2018predicting,chen2020pathomic,chen2022pan,ding2023pathology,jaume2024modeling,xu2023multimodal}.
% Among these, attention-based fusion mechanisms~\cite{chen2020pathomic,chen2021multimodal,jaume2024modeling,song2024multimodal} have emerged as the dominant paradigm, effectively modeling cross-modal correlations~\cite{xu2023multimodal} and enabling semantically coherent integration.

A common assumption behind these models is that aligning heterogeneous modalities in a shared latent space will improve prediction. However, recent evidence suggests that this assumption does not always hold~\cite{zhang2024prototypical,qu2025multimodal,xu2025distilled}. 
When alignment is enforced too aggressively, multimodal representations can lose modality-specific structure. 
This phenomenon, which we refer to as \emph{over-alignment collapse}, occurs when heterogeneous signals are prematurely forced into a shared representation space. 
In practice, this may dilute morphology-rich spatial patterns in WSIs and gene-level variation in genomics, reducing representational diversity and limiting the benefit of multimodal integration. 
Empirical observations further reveal a surprising outcome: multimodal models can sometimes underperform strong WSI-only baselines~\cite{ilse2018attention,lin2023interventional,shao2021transmil,ren2025otsurv}, which naturally preserve detailed morphological information.

% Despite these advances, recent studies~\cite{zhang2024prototypical,qu2025multimodal,xu2025distilled} uncover a fundamental challenge: when heterogeneous modalities are prematurely forced into a shared latent space, the model tends to \textbf{over-align} their representations.
% We refer to this as \textit{over-alignment collapse}: the degradation where aggressive cross-modal alignment erases modality-specific structure, diluting morphology-rich spatial cues in WSIs and gene-level variations in genomics, thereby diminishing representational diversity and model robustness.
% Empirically, this phenomenon manifests as a counterintuitive trend: multi-modal models can underperform compared to single-modality WSI baselines~\cite{ilse2018attention,lin2023interventional,shao2021transmil,ren2025otsurv}, which inherently preserve fine-grained morphological information.

Several recent works attempt to mitigate this issue. 
PIBD~\cite{zhang2024prototypical} disentangles shared and modality-specific representations, while MRePath~\cite{qu2025multimodal} dynamically reweights modality contributions. 
Although these approaches partially alleviate redundancy, they do not fully resolve the tension between cross-modal alignment and modality-specific information preservation. 
In particular, most existing methods treat these objectives independently, lacking a unified principle that jointly encourages semantic agreement across modalities while maintaining the distinctive signals each modality provides.

In this work, we revisit multimodal survival modeling from this perspective. 
Our key observation is simple: effective multimodal representations should \emph{first discover what modalities share, and only then preserve what makes them different}. We term this learning principle \textbf{Together-Then-Apart}.  
Following this idea, we propose \textbf{TTA}, a framework that explicitly balances cross-modal alignment and modality-specific distinctiveness.
In the \textit{Together} stage, TTA aligns modalities by projecting both WSI and genomic embeddings onto a shared set of prototypes that serve as semantic anchors.
Unlike prior approaches that maintain separate prototype spaces for each modality~\cite{zhang2024prototypical,song2024multimodal}, our shared-prototype design captures cross-modal structure while avoiding premature homogenization. 
To further accommodate intra-sample heterogeneity, we introduce an \textbf{unbalanced optimal transport (UOT)} formulation that learns entropy-regularized instance-to-prototype assignments under relaxed marginal constraints. 
This allows the model to selectively emphasize informative regions and produce stable MIL representations.
In the \textit{Apart} stage, TTA preserves modality-specific information. 
We introduce modality-specific anchors that retain distinctive semantics for WSIs and genomics. 
A contrastive regularization encourages features to remain close to their modality anchors while maintaining separation across modalities, preventing representational collapse and complementing the alignment learned in the \textit{Together} stage. 
Through this alternating process, TTA produces multimodal representations that capture both shared prognostic signals and modality-specific evidence.
Our contributions are as follows:

\begin{itemize}
    % \item We identify \emph{over-alignment collapse} as a key challenge in multimodal survival analysis and highlight the importance of balancing cross-modal alignment with modality-specific distinctiveness.
    % \item We propose \textbf{Together-Then-Apart (TTA)}, a unified framework that combines shared-prototype alignment with anchor-based contrastive learning, guided by an unbalanced optimal transport formulation.
    % \item Extensive experiments on five TCGA cohorts demonstrate that TTA consistently improves survival prediction over recent multimodal survival models while providing interpretable cross-modal structures.
\item We identify \textit{over-alignment collapse} as a key challenge in multimodal survival analysis, where aggressive cross-modal alignment suppresses modality-specific signals and limits the benefit of multimodal integration.
\item We propose \textbf{Together-Then-Apart (TTA)}, a framework that balances cross-modal alignment and modality-specific distinctiveness. TTA aligns modalities through shared prototypes while preserving distinctive information using modality-specific anchors and contrastive regularization.
\item Extensive experiments on five TCGA cohorts demonstrate that TTA consistently improves survival prediction over recent multimodal survival models and reveals interpretable cross-modal structures related to clinical outcomes.
\end{itemize}

\section{Related Works}
\label{sec:related_works}

\subsection{Single-Modality Survival Analysis}
Recent studies have explored survival prediction using a single data modality, most commonly histopathology whole-slide images (WSIs) or genomic profiles. 
Due to the gigapixel resolution of WSIs, most approaches formulate the task within a Multiple Instance Learning (MIL) framework, where a slide is represented as a bag of image patches. 
A large body of work focuses on how to aggregate patch-level features into slide-level representations. 
Early methods employ attention-based aggregation to assign importance weights to instances before pooling~\cite{ilse2018attention,lu2021data}. 
Subsequent works extend this paradigm by modeling contextual dependencies among patches using sequence-based architectures~\cite{shao2021transmil,yang2024mambamil}. 
Other approaches exploit the hierarchical organization of WSI patches through multi-scale or hierarchical frameworks~\cite{chen2022scaling,shao2023hvtsurv}. 
Graph-based methods have also been proposed to model spatial relationships between tissue regions and capture context-aware interactions~\cite{chen2021whole}. 
Another line of research focuses on reducing redundancy among patch tokens through prototype-based abstraction~\cite{yao2020whole,VU2023handcrafted,song2024morphological,claudio2024mapping} or instance filtering mechanisms that select informative regions for downstream prediction~\cite{lin2023interventional,li2021dual,zhang2022dtfd,ren2025otsurv,zhang2025supervise}.
In parallel, genomic survival modeling typically relies on feedforward neural networks or self-normalizing neural architectures to process transcriptomic or pathway-level features~\cite{haykin1998neural,klambauer2017self}. 
Although these single-modality approaches provide strong predictive baselines and robust modality-specific representations, they capture only one aspect of tumor biology. 
Integrating complementary modalities therefore becomes a natural direction for improving survival prediction.

\subsection{Multimodal Survival Analysis}
Recent research efforts increasingly focus on integrating heterogeneous modalities~\cite{wen2025beyond,guo2026csrv2,you2024calibrating,you2025uncovering}, particularly histopathology and genomics, for prognostic modeling. 
Many multimodal methods adopt attention-based mechanisms to capture interactions between modalities and learn joint representations for survival prediction~\cite{chen2021multimodal,zhou2023cross,jaume2024modeling}. 
Beyond attention-based approaches, optimal transport has also been explored as a mechanism for aligning pathology and genomic representations in a shared space~\cite{xu2023multimodal,song2024multimodal}. 
To improve scalability, MMP~\cite{song2024multimodal} introduces morphological prototypes to compress redundant WSI tokens. 
PIBD~\cite{zhang2024prototypical} adopts an information-theoretic framework to disentangle shared and modality-specific representations and reduce redundancy. 
MRePath~\cite{qu2025multimodal} further introduces dynamic reweighting to rebalance modality contributions. 
Other studies address robustness in the presence of missing modalities, including LD-CVAE~\cite{zhou2025robust} and DisPro~\cite{xu2025distilled}. 
Despite these advances, most existing approaches focus primarily on modeling cross-modal correlations through alignment. 
How to capture shared structures across modalities while preserving modality-specific signals remains an open challenge in multimodal survival analysis.

% The integration of heterogeneous modalities, especially histopathology and genomics, has become central to improving prognostic modeling. Recent multi-modal approaches commonly employ attention-based fusion mechanisms~\cite{chen2021multimodal,zhou2023cross,jaume2024modeling} to capture fine-grained cross-modal dependencies.
% Beyond attention-based fusion, optimal transport has also been explored as a tool for cross-modal alignment between pathology and genomics representations~\cite{xu2023multimodal,song2024multimodal}.
% To address scalability, MMP~\cite{song2024multimodal} introduces morphological prototypes that compress redundant WSI tokens.
% PIBD~\cite{zhang2024prototypical} adopts an information-theoretic perspective to reduce both inter- and intra-modal redundancy, while MRePath~\cite{qu2025multimodal} dynamically rebalances modality contributions to mitigate imbalance.
% Beyond alignment, LD-CVAE~\cite{zhou2025robust} and DisPro~\cite{xu2025distilled} improve robustness under missing-modality conditions.
% Despite these advances, most methods focus primarily on \textit{alignment}. A comprehensive framework that jointly models cross-modal correlations while explicitly maintaining modality-specific distinctiveness remains largely unexplored.

\subsection{Survival Analysis with Optimal Transport}
Optimal Transport (OT) has recently emerged as an effective tool for modeling structural relationships and heterogeneity in survival data. 
In WSI analysis, unbalanced OT has been used for imbalanced clustering and instance-level patch selection, allowing models to focus on salient regions that drive prognosis~\cite{ren2025otsurv}. 
A similar idea can be applied to genomic features, where transporting mass toward informative pathway tokens yields sparse and biologically meaningful representations.
Beyond within-modality modeling, OT has also been explored for cross-modal alignment between pathology and genomic representations~\cite{xu2023multimodal,song2024multimodal}. 
Existing OT-based multimodal approaches typically compute transport plans either between modality-specific embeddings~\cite{xu2023multimodal} or between modality-specific prototype assignments~\cite{song2024multimodal}. 
In contrast, our method computes a single transport plan after concatenating pathology and genomic tokens and transporting them to a shared prototype bank. 
This formulation enables cross-modal evidence to interact under a unified assignment geometry while maintaining flexibility through an unbalanced OT formulation.

\section{Method}
\label{sec:method}
\label{sec:method_overview}

\begin{figure*}[t]
\centering
\includegraphics[width=0.99\linewidth]{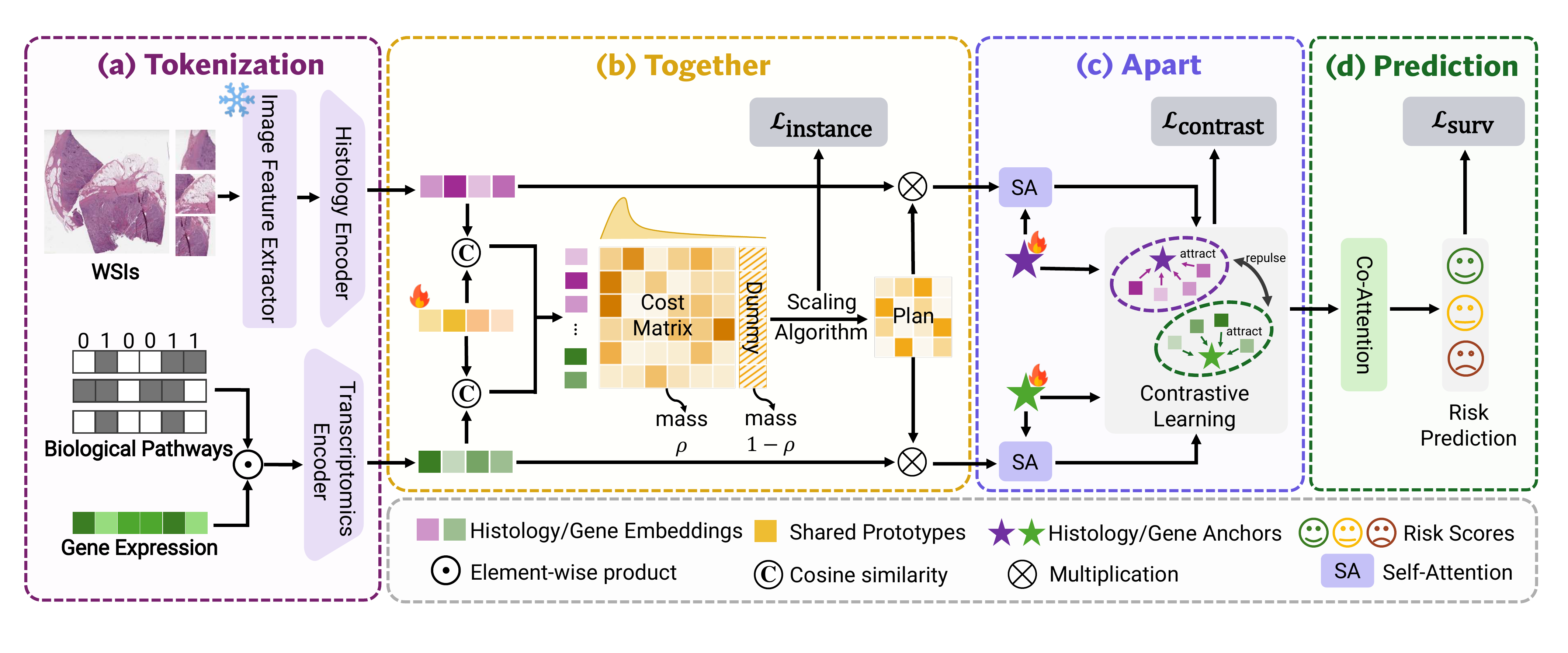}
\caption{\textbf{Overview of TTA.} 
(1) \textit{Tokenization:} WSIs and gene-expression profiles are converted into modality-specific token sets.
(2) \textit{\textsc{Together}:} All tokens are assigned to a shared prototype bank via a semi-relaxed unbalanced optimal transport (UOT) module with a curriculum on the mass parameter~$\rho$.
(3) \textit{\textsc{Apart}:} Modality representations are refined with modality-specific anchors and a contrastive regularizer to retain modality identity.
(4) \textit{Prediction:} A co-attention module exchanges information between modalities and outputs the survival risk.}
\label{fig:overview}
\end{figure*}

We consider survival prediction from two complementary patient views: pathology WSIs and transcriptomics. 
A common practice is to tightly align the two modalities, but in survival analysis this can be brittle: WSIs contain many ambiguous patches, and transcriptomic signals are sparse and patient-dependent. 
If the model is forced to agree everywhere, modality-specific cues can be washed out, and the combined model may even fall behind strong single-modality baselines.
TTA follows a simple principle: \emph{align where the modalities agree, and keep them separated where they should differ}. 
As shown in Fig.~\ref{fig:overview}, TTA consists of two stages. 
\textsc{Together} (Sec.~\ref{sec:method_together_stage}) aligns both modalities through shared prototypes using an unbalanced optimal transport assignment, designed to tolerate uncertainty and long-tailed heterogeneity. 
\textsc{Apart} (Sec.~\ref{sec:method_apart_stage}) then re-introduces modality identity using modality-specific anchors and a lightweight contrastive objective. 
Finally, a co-attention prediction head aggregates complementary evidence for survival risk estimation (Sec.~\ref{sec:method_fusion_and_prediction}).

\subsection{Problem Formulation}
\label{sec:method_formulation}
% To map both modalities from raw observations into compact, modality-appropriate feature embeddings, we follow the common practice \cite{song2024multimodal}. We first introduce how raw WSIs and transcriptomics are converted into token sets that are well-posed for subsequent \textit{alignment} and \textit{diversification}.
We represent each patient with two token sets, one per modality, so that alignment and refinement can be performed at the token level~\cite{song2024multimodal}.

\myparagraph{WSI tokenization.}
Each WSI is divided into patches and encoded by a pretrained image encoder such as ResNet50 or UNI~\cite{chen2024uni}. 
A linear projection maps patch embeddings to $D$ dimensions, yielding pathology tokens
$\boldsymbol{X}^p_n \in \mathbb{R}^{N^p_n\times D}$ for patient $n$, where $N^p_n$ is the number of patches and $\boldsymbol{x}^p_{n,i}$ denotes the $i$-th patch token.

\myparagraph{Gene-expression tokenization.}
We formulate transcriptomics into pathway-level tokens to obtain compact and interpretable genomic factors. 
Following MMP~\cite{song2024multimodal}, we use a fixed pathway collection indexed by $c\in\{1,\ldots,C_g\}$.
Each pathway is specified by a binary selector $\boldsymbol{a}_{c}\!\in\!\{0,1\}^{G}$ over $G$ genes. 
Given the gene-expression vector $\boldsymbol{x}^g_{n}\!\in\!\mathbb{R}^{G}$, we extract the pathway-specific subprofile and densify it:
\begin{equation}
\label{eq:pathway-select}
\boldsymbol{z}^{\mathrm{agg}}_{n,c}=R(\boldsymbol{x}^g_{n} \odot \boldsymbol{a}_{c})\in\mathbb{R}^{N_{c}},
\end{equation}
where $\odot$ denotes element-wise multiplication, $R(\cdot)$ removes zeros, and $N_c$ is the number of genes selected by pathway $c$. 
A lightweight per-pathway head then maps each summary to a $D$-dimensional token:
\begin{equation}
\label{eq:pathway-embed}
\boldsymbol{x}^g_{n,c}=E_g\!\big(\boldsymbol{z}^{\mathrm{agg}}_{n,c}\big)\in\mathbb{R}^{D}.
\end{equation}
This produces genomic tokens $\boldsymbol{X}^g_n=[\boldsymbol{x}^g_{n,1};\ldots;\boldsymbol{x}^g_{n,C_g}]\in\mathbb{R}^{N^g_n\times D}$ with $N^g_n=C_g$. 
In our experiments, pathways (e.g., Hallmark~\cite{liberzon2015molecular}) are fixed a priori.

\subsection{\textbf{\textsc{Together}}: UOT-guided Prototype Alignment}
\label{sec:method_together_stage}
The \textsc{Together} stage aligns histology and genomics using a shared set of prototypes. 
Instead of directly matching tokens across modalities, we map each token to the same prototype bank, which serves as a common coordinate system. 
Crucially, assignments are computed with unbalanced optimal transport, allowing uncertain tokens to remain diffuse early on and preventing noisy tokens from dominating alignment.

\myparagraph{Shared prototypes.}
Let $P\!\in\!\mathbb{R}^{K\times D'}$ be $K$ learnable prototypes in a shared $D'$-dimensional space. 
For modality $m\!\in\!\{p,g\}$, tokens are projected by $W_m$ and $\ell_2$-normalized to obtain $\tilde X^m_n\!\in\!\mathbb{R}^{N^m_n\times D'}$. 
Prototypes are also normalized to $\tilde P\!\in\!\mathbb{R}^{K\times D'}$. 
We compute cosine similarities with temperature $\tau$:
\begin{equation}
L^m_n=\frac{\tilde X^m_n (\tilde P)^\top}{\tau}\in\mathbb{R}^{N^m_n\times K}.
\end{equation}

\myparagraph{Unbalanced optimal transport for robust assignments.}
A naive choice is to use $\mathrm{softmax}(L^m_n)$ as token-to-prototype weights. 
In survival analysis, this can be unstable: WSI bags are large and long-tailed, and pathway tokens exhibit strong patient-specific sparsity. 
Moreover, both modalities contain noisy or weakly informative tokens, especially early in training. 
Motivated by OTSurv~\cite{ren2025otsurv}, we compute assignments via unbalanced optimal transport (UOT), which naturally supports \emph{semi-relaxed} matching: each token can spread its mass across prototypes rather than committing prematurely.
We define a transport cost using negative log-probabilities:
\begin{equation}
C^m_n= -\log \mathrm{softmax}(L^m_n)\in\mathbb{R}^{N^m_n\times K},\quad m\in\{p,g\}.
\end{equation}
To allow cross-modal evidence sharing under a single assignment geometry, we concatenate the two modalities along the token axis:
\begin{equation}
\label{eq:cost}
C_n = \begin{bmatrix} C^p_n \\ C^g_n \end{bmatrix} \in \mathbb{R}^{N_{\mathrm{tot}}\times K}, \quad N_{\mathrm{tot}} = N^p_n + N^g_n.
\end{equation}
We seek a transport plan $Q\!\in\!\mathbb{R}^{N_{\mathrm{tot}}\times K}_{\ge 0}$ that minimizes a general OT objective:
\begin{equation}
\label{eq:gen}
\min_{Q\ge 0}\;\; \langle Q, C_n\rangle_F \;+\; F_1\!\big(Q\mathbf{1},\, u\big) \;+\; F_2\!\big(Q^\top\mathbf{1},\, v\big),
\end{equation}
where $\langle\cdot,\cdot\rangle_F$ denotes the Frobenius inner product, $\mathbf{1}$ is an all-one vector with context-dependent dimension, $u$ and $v$ are source and target marginals, and $F_1,F_2$ encode the corresponding marginal constraints. 
We fix the source as uniform $u=a=\tfrac{1}{N_{\mathrm{tot}}}\mathbf{1}_{N_{\mathrm{tot}}}$, ensuring each token contributes equally before assignment. 
The prototype-side marginal is relaxed to handle heterogeneity.

\myparagraph{Prototype-side relaxation.}
We penalize deviation from a uniform prior $b=\tfrac{1}{K}\mathbf{1}_K$ via KL divergence:
\begin{equation}
\label{eq:new_uot}
\min_{Q\ge 0}\quad \langle Q, C_n\rangle \;+\; \gamma\, \mathrm{KL}\!\big(Q^\top \mathbf{1}\,\big\|\, b\big) \quad \text{s.t.} \quad Q\,\mathbf{1}=a,
\end{equation}
where $\gamma\!>\!0$ controls the relaxation strength. 
This prevents the plan from collapsing onto a few dominant prototypes and keeps prototype usage balanced, while still allowing patient-dependent deviations. 
We provide additional algorithm details and theoretical  derivations in Appendix~\ref{sec:tta_theory}.
% This prevents the plan from collapsing onto a few dominant prototypes and keeps prototype usage balanced, while still allowing patient-dependent deviations. 
% We provide additional algorithm details and theoretical  derivations in Appendix~A.

\myparagraph{Curriculum on transported mass.}
Even with a balanced prior, noisy tokens can still introduce spurious matches early in training. 
We therefore control the \emph{total mass} transported to real prototypes using $\rho\!\in\![0,1]$: only a $\rho$-fraction of mass is assigned to the $K$ prototypes, and the remaining $1-\rho$ is absorbed by a sink. 
Small $\rho$ yields conservative, uncertainty-tolerant assignments; larger $\rho$ gradually enforces more decisive matching.
Let $b_\rho=\tfrac{\rho}{K}\mathbf{1}_{K}$. We obtain:
\begin{equation}
\label{eq:rho_ot}
\min_{Q\ge 0}\quad \langle Q, C_n\rangle \;+\; \gamma\, \mathrm{KL}\!\big(Q^\top \mathbf{1}\,\big\|\, b_\rho\big) \quad \text{s.t.} \quad Q\,\mathbf{1}\le a,\; \mathbf{1}^{\top}Q\mathbf{1}=\rho.
\end{equation}
We schedule $\rho$ with a smooth ramp-up:
\begin{equation}
\label{eq:rho}
\rho(t)=\rho_{\mathrm{base}} + (\rho_{\mathrm{upper}}\!-\!\rho_{\mathrm{base}})\,
\exp\!\left(-5\Big(1-\tfrac{t}{T}\Big)^2\right),
\end{equation}
where $t$ is the current training iteration clipped to $[0,T]$, $T$ is the ramp-up horizon, and $\rho_{\mathrm{base}}$ and $\rho_{\mathrm{upper}}$ denote the initial and maximum transported masses, respectively.
Thus, assignments remain cautious during the unstable early phase and become sharper only after representations have matured.

To solve the relaxed problem efficiently, we append a zero-cost sink column and define $\tilde C_n=[C_n\;|\;\mathbf{0}]\in\mathbb{R}^{N_{\mathrm{tot}}\times(K+1)}$. 
The corresponding target prior becomes:
\begin{equation}
\tilde b(\rho)=\begin{bmatrix}\tfrac{\rho}{K}\mathbf{1}_{K}\\[2pt] 1-\rho\end{bmatrix}\in\mathbb{R}^{K+1}.
\end{equation}
We then solve:
\begin{equation}
\label{eq:uot}
\min_{\tilde Q\ge 0}\quad \langle \tilde Q,\tilde C_n\rangle \;+\; \gamma\, \mathrm{KL}\!\left(\tilde Q^\top \mathbf{1}\; \middle\| \; \tilde b(\rho)\right) \quad
\text{s.t.} \quad \tilde Q\,\mathbf{1}=a,
\end{equation}
After removing the sink column, we multiply the remaining plan by $N_{\mathrm{tot}}$ and denote the resulting row-scaled UOT assignment by $Q^\star\!\in\!\mathbb{R}^{N_{\mathrm{tot}}\times K}$. 
Because the plan is computed over concatenated tokens, strong evidence from one modality can support the other, while the sink prevents forced agreement when the signals conflict.
Let $Q^{\star,p}$ and $Q^{\star,g}$ denote the modality-specific blocks of $Q^\star$. 
We blend similarity-based weights and transport-based assignments, then row-normalize the mixture over prototypes:
\begin{equation}
W^m_n=\mathrm{Normalize}\!\left((1-\beta)\,\mathrm{softmax}(L^m_n)+\beta\,Q^{\star,m}\right),\quad \beta\in[0,1],
\end{equation}
where $\mathrm{Normalize}(A)_{ik}=A_{ik}/(\sum_j A_{ij}+\epsilon)$ denotes row-wise normalization with a small $\epsilon>0$ for numerical stability.
Prototype tokens are aggregated as:
\begin{equation}
\label{eq:agg}
H^m_n = (W^m_n)^\top X^m_n \in \mathbb{R}^{K\times D},\quad m\in\{p,g\}.
\end{equation}

\myparagraph{Instance-level supervision.}
We further use UOT assignments as soft pseudo-labels to regularize token-to-prototype predictions. 
Let $\boldsymbol{\pi}^m_{i}=\mathrm{Normalize}(Q^{\star,m})_{i,:}$ be the row-normalized UOT pseudo-label and $\boldsymbol{p}^m_{i}=\mathrm{softmax}(\boldsymbol{\ell}^m_{i})$ be the predicted prototype distribution, where $\boldsymbol{\ell}^m_i$ is the $i$-th row of $L^m_n$. 
We define:
\begin{equation}
\mathcal{L}^m_{\mathrm{CE}}
= -\frac{1}{N^m_n}\sum_{i=1}^{N^m_n}\!\left\langle \boldsymbol{\pi}^m_{i},\, \log \boldsymbol{p}^m_{i} \right\rangle,\quad m\in\{p,g\},
\end{equation}
and combine the two modalities:
\begin{equation}
\label{eq:instance}
\mathcal{L}_{\mathrm{instance}}
= \lambda_{\mathrm{wsi}}\,\mathcal{L}^{p}_{\mathrm{CE}}
 + \lambda_{\mathrm{gen}}\,\mathcal{L}^{g}_{\mathrm{CE}}.
\end{equation}

\subsection{\textbf{\textsc{Apart}}: Contrastive Diversification}
\label{sec:method_apart_stage}
After \textsc{Together}, both modalities live in the same prototype coordinate system. 
This is useful for sharing information, but it can also blur modality identity. 
The \textsc{Apart} stage counteracts this effect by explicitly keeping histology and genomics distinguishable.

\myparagraph{Anchor refinement.}
We introduce a learnable anchor for each modality, $a^{p}$ and $a^{g}\!\in\!\mathbb{R}^{D'}$, which serve as modality identifiers rather than patient-specific centers. 
Prototype tokens are projected by a modality-specific map $U_m\!\in\!\mathbb{R}^{D\times D'}$ and normalized to obtain $\tilde H^m_n\!\in\!\mathbb{R}^{K\times D'}$. 
We append the anchor as an extra token and refine the set with a lightweight self-attention module $\mathcal{R}_\theta$:
\begin{equation}
Y^m_n=\mathcal{R}_\theta([\tilde H^m_n;\, a^m])\in\mathbb{R}^{(K+1)\times D'},\quad m\in\{p,g\},
\end{equation}
then keep the first $K$ tokens:
\begin{equation}
\hat H^m_n=\mathrm{head}_{K}(Y^m_n)\in\mathbb{R}^{K\times D'}.
\end{equation}

\myparagraph{Anchor-based contrastive regularization.}
We encourage each modality to stay close to its own anchor while remaining separated from the other modality's anchor. 
Let $\phi(\cdot)$ be a projection head followed by unit normalization, and let $\bar h^m_n=\tfrac{1}{K}\sum_{k=1}^{K} \hat h^m_{n,k}$ be the mean refined token. 
We compute:
\begin{equation}
s^m_{+}=\frac{\langle \phi(\bar h^m_n),\phi(a^m)\rangle}{\tau_r},\quad
s^m_{-}=\frac{\langle \phi(\bar h^m_n),\phi(a^{\bar m})\rangle}{\tau_r}, \quad m\in\{p,g\},
\end{equation}
where $\bar m$ denotes the other modality and $\tau_r>0$ is the contrastive temperature.
We then define the InfoNCE-style loss~\cite{oord2018representation,chen2020graph}:
\begin{equation}
\label{eq:contrast}
\begin{aligned}
\mathcal{L}_{\mathrm{contrast}}
= -\log\frac{\exp(s^{p}_{+})}{\exp(s^{p}_{+})+\exp(s^{p}_{-})} 
   -\log\frac{\exp(s^{g}_{+})}{\exp(s^{g}_{+})+\exp(s^{g}_{-})}.
\end{aligned}
\end{equation}

\subsection{Prediction Head}
\label{sec:method_fusion_and_prediction}
We next combine the two refined modality streams for survival prediction. 
We use a transformer-style co-attention module~\cite{vaswani2017attention} over prototype tokens $\hat H^p_n$ and $\hat H^g_n$. 
Co-attention allows pathology to query genomics and vice versa, capturing cross-modal dependencies at the prototype level. 
We then mean-pool within each modality to obtain $h^p_n$ and $h^g_n$, and map them to a joint representation $f_n = F(h^p_n,h^g_n)\in\mathbb{R}^{d_f}$. 
A linear risk head predicts log-risk $r_n = \langle w, f_n \rangle$, with $w\in\mathbb{R}^{d_f}$.
We train with a standard survival objective $\mathcal{L}_{\mathrm{surv}}$ (e.g., Cox partial likelihood~\cite{cox1972regression}). 
The full training loss is:
\begin{equation}
\label{eq:total_loss}
\mathcal{L}_{\mathrm{total}}=\mathcal{L}_{\mathrm{surv}}
 +\lambda_{\mathrm{contrast}}\mathcal{L}_{\mathrm{contrast}}
 +\lambda_{\mathrm{inst}}\mathcal{L}_{\mathrm{instance}}.
\end{equation}

\myparagraph{Discussion.}
The two auxiliary terms play different roles. 
$\mathcal{L}_{\mathrm{instance}}$ stabilizes prototype assignments in \textsc{Together}, while $\mathcal{L}_{\mathrm{contrast}}$ preserves modality identity in \textsc{Apart}. 
In practice, we find this separation improves both robustness and interpretability without complicating optimization (see ablation studies in Section~\ref{subsec:sblation} and Appendix~\ref{sec:balancing_perspective}).
% In practice, we find this separation improves both robustness and interpretability without complicating optimization (see ablation studies in Section~\ref{subsec:sblation} and Appendix~A.1).

\section{Experiments}
\label{sec:experiment}
In this section, we first describe the implementation details (Section~\ref{subsec:implementation}). 
We then conduct comprehensive comparisons with seventeen state-of-the-art methods across five benchmarks (Section~\ref{subsec:mainresults}). We further perform ablation studies to analyze the contributions of the main components of TTA (Section~\ref{subsec:sblation}).

\subsection{Implementation Details}
\label{subsec:implementation}
\myparagraph{Datasets.} 
We evaluate our method on five cancer cohorts from The Cancer Genome Atlas (TCGA): Bladder urothelial carcinoma (BLCA, $n=359$), Breast invasive carcinoma (BRCA, $n=868$), Stomach adenocarcinoma (STAD, $n=318$), Colon and Rectum adenocarcinoma (CRC, $n=296$), and Kidney renal clear cell carcinoma (KIRC, $n=340$). 
These cohorts span distinct tissue systems, including gastrointestinal (CRC, STAD), urological (BLCA, KIRC), and breast (BRCA) cancers, and exhibit substantially different histology, genomic profiles, and survival patterns. 
This diversity provides a challenging testbed for evaluating cross-tissue generalization. 

We follow the data splits in~\cite{song2024multimodal} and use disease-specific survival (DSS)~\cite{liu2018integrated} as the prediction target. 
Performance is evaluated using the concordance index (C-Index) with 5-fold site-stratified cross-validation~\cite{howard2021impact}, which is widely adopted for survival analysis on TCGA cohorts with limited sample sizes~\cite{song2024multimodal,chen2021multimodal,xu2023multimodal,jaume2024modeling}. 
To avoid overfitting, a single configuration is used for all folds and datasets without per-fold tuning, and we report mean$\pm$std C-Index.

\myparagraph{Settings.} 
Key configurations are summarized here. Additional implementation details are provided in {Appendix~\ref{sec:tta_imple_details}}, and further experiments are reported in {Appendix~\ref{sec:tta_addtional_experiments}}.
% Key configurations are summarized here. Additional implementation details are provided in {Appendix~B}, and further experiments are reported in {Appendix~C}.

WSI patches are encoded using UNI~\cite{chen2024towards}, a DINOv2-based ViT-Large pretrained on $1{\times}10^8$ pathology patches. 
To verify robustness across feature extractors, we also evaluate ResNet50~\cite{deng2009imagenet} (Table~\ref{tab:extractor_resnet50} in Appendix~\ref{sec:tta_addtional_experiments}).
% To verify robustness across feature extractors, we also evaluate ResNet50~\cite{deng2009imagenet} (Table~9 in Appendix~C). 
For bags exceeding a fixed size, we apply uniform subsampling. Shorter bags are zero-padded. 
The UOT transport plan then learns importance weights during aggregation. 
Following MMP~\cite{song2024multimodal}, gene expression profiles are organized into $50$ Hallmark pathways covering $4{,}241$ genes, which correspond to well-defined biological processes and provide interpretable genomic groupings~\cite{liberzon2015molecular}. 
In the \textsc{Together} stage, we use $K{=}32$ shared prototypes with KL regularization weight $\gamma{=}0.1$. 
The curriculum mass is scheduled from $\rho_{\mathrm{base}}{=}0.1$ to $\rho_{\mathrm{upper}}{=}1.0$. 
The auxiliary loss weights are $\lambda_{\mathrm{contrast}}{=}\lambda_{\mathrm{inst}}{=}0.5$, and the modality weights are $\lambda_{\mathrm{wsi}}{=}\lambda_{\mathrm{gen}}{=}1$. 
For the survival objective $\mathcal{L}_{\mathrm{surv}}$, we use the Cox partial likelihood~\cite{cox1972regression}. 
Both Cox and discrete-time NLL are standard objectives in survival modeling, and we verify their interchangeability in Appendix~\ref{sec:tta_addtional_experiments}.
% Both Cox and discrete-time NLL are standard objectives in survival modeling, and we verify their interchangeability in Appendix~C.
Note that Cox partial likelihood requires batch sizes larger than $1$, as risk set computation involves pairwise comparisons between samples.

\begin{table*}[t]
\centering
\caption{\textbf{Comparison with state-of-the-art methods} in terms of C-index (mean $\pm$ std) across five TCGA cancer cohorts. ``Modal.'' denotes Modality, ``h.'' and ``g.'' denote models that rely solely on histopathology WSIs and genomic data, respectively, while ``g.+h.'' indicates multimodal models using both modalities. 
The overall score is computed as the average C-index across datasets. 
The best and second-best results are highlighted in \textbf{bold} and \underline{underline}, respectively.}
\label{table:table1}
\scriptsize
\setlength{\tabcolsep}{1.05pt}
\renewcommand{\arraystretch}{1.00}
\begin{tabular}{c|c|cccccc}
\toprule
{\textbf{Model}} & {\textbf{Modal.}} & \textbf{BRCA} & \textbf{BLCA} & \textbf{STAD} & \textbf{CRC} & \textbf{KIRC} & {\textbf{Overall}} \\
\midrule
SNN~\cite{klambauer2017self} & g. &
\pmstd{0.619}{0.063} & \pmstd{0.618}{0.027} & \pmstd{0.557}{0.072} & \pmstd{0.576}{0.102} & \pmstd{0.689}{0.098} & 0.611 \\
SNNTrans\cite{shao2021transmil} & g. &
\pmstd{0.630}{0.062} & \pmstd{0.622}{0.038} & \pmstd{0.559}{0.070} & \pmstd{0.570}{0.098} & \pmstd{0.698}{0.126} & 0.616 \\
\midrule
ABMIL~\cite{ilse2018attention} & h. &
\pmstd{0.566}{0.097} & \pmstd{0.553}{0.052} & \pmstd{0.566}{0.033} & \pmstd{0.655}{0.118} & \pmstd{0.675}{0.124} & 0.603 \\
CLAM~\cite{lu2021data} & h. &
\pmstd{0.627}{0.188} & \pmstd{0.618}{0.104} & \pmstd{0.538}{0.077} & \pmstd{0.629}{0.132} & \pmstd{0.670}{0.124} & 0.616 \\
HIPT~\cite{chen2022scaling} & h. &
\pmstd{0.566}{0.092} & \pmstd{0.606}{0.126} & \pmstd{0.529}{0.074} & \pmstd{0.600}{0.066} & \pmstd{0.685}{0.099} & 0.597 \\
AttnMISL~\cite{yao2020whole} & h. &
\pmstd{0.585}{0.073} & \pmstd{0.533}{0.065} & \pmstd{0.541}{0.052} & \textbf{\pmstd{0.723}{0.111}} & \pmstd{0.656}{0.112} & 0.607 \\
TransMIL~\cite{shao2021transmil} & h. &
\pmstd{0.599}{0.056} & \pmstd{0.595}{0.102} & \pmstd{0.500}{0.060} & \pmstd{0.550}{0.143} & \pmstd{0.676}{0.132} & 0.584\\
OTSurv~\cite{ren2025otsurv} & h. &
\pmstd{0.625}{0.071} & \pmstd{0.637}{0.065} & \pmstd{0.556}{0.057} & \pmstd{0.663}{0.102} & \pmstd{0.739}{0.149} & 0.644 \\
PANTHER~\cite{song2024morphological} & h. &
\pmstd{0.643}{0.128} & \pmstd{0.593}{0.083} & \pmstd{0.503}{0.082} & \pmstd{0.639}{0.133} & \pmstd{0.700}{0.124} & 0.615 \\
\midrule
MCAT~\cite{chen2021multimodal} & g.+ h. &
\pmstd{0.650}{0.096} & \pmstd{0.623}{0.055} & \pmstd{0.528}{0.112} & \pmstd{0.579}{0.134} & \pmstd{0.699}{0.121} & 0.615 \\
CMTA~\cite{zhou2023cross} & g.+ h. &
\pmstd{0.687}{0.077} & \pmstd{0.622}{0.056} & \pmstd{0.539}{0.076} & \pmstd{0.568}{0.102} & \pmstd{0.711}{0.121} & 0.625 \\
MOTCat~\cite{xu2023multimodal} & g.+ h. &
\pmstd{0.717}{0.058} & \pmstd{0.631}{0.059} & \pmstd{0.576}{0.075} & \pmstd{0.598}{0.128} & \pmstd{0.708}{0.109} & 0.646 \\
SurvPath~\cite{jaume2024modeling} & g.+ h. &
\pmstd{0.696}{0.061} & \pmstd{0.619}{0.051} & \pmstd{0.555}{0.133} & \pmstd{0.588}{0.134} & \pmstd{0.742}{0.101} & 0.640 \\
PIBD~\cite{zhang2024prototypical} & g.+ h. &
\pmstd{0.683}{0.056} & \pmstd{0.661}{0.034} & \pmstd{0.552}{0.054} & \pmstd{0.582}{0.077} & \pmstd{0.732}{0.139} & 0.642 \\
MRePath~\cite{qu2025multimodal} & g.+ h. &
\pmstd{0.703}{0.036} & \pmstd{0.651}{0.060} & \pmstd{0.579}{0.062} & \pmstd{0.639}{0.101} & \pmstd{0.743}{0.112} & 0.663 \\
MMP~\cite{song2024multimodal} & g.+ h. &
\underline{\pmstd{0.724}{0.067}} & \pmstd{0.640}{0.050} & \underline{\pmstd{0.594}{0.066}} & \pmstd{0.634}{0.118} & \pmstd{0.743}{0.133} & \underline{0.667} \\
LD-CVAE~\cite{zhou2025robust} & g.+ h. &
\pmstd{0.709}{0.047} & \textbf{\pmstd{0.676}{0.035}} & \pmstd{0.589}{0.083} & \pmstd{0.602}{0.120} & \underline{\pmstd{0.751}{0.139}} & 0.665 \\
\rowcolor{gray!20}
\textbf{TTA}(Ours) & g.+ h. &
\textbf{\pmstd{0.726}{0.039}} & \underline{\pmstd{0.662}{0.079}} & \textbf{\pmstd{0.613}{0.079}} &
\underline{\pmstd{0.685}{0.131}} & \textbf{\pmstd{0.778}{0.117}} & \textbf{0.693} \\
\bottomrule
\end{tabular}
\end{table*}

\subsection{Main Results}
\label{subsec:mainresults}
\myparagraph{Comparisons with SOTAs.}
We compare TTA with a wide range of recent survival analysis methods. \textit{Unimodal baselines}, including SNN~\cite{klambauer2017self} and SNNTrans~\cite{klambauer2017self, shao2021transmil} for genomic survival prediction. For histopathology-only survival modeling, we compare against seven representative MIL methods: ABMIL~\cite{ilse2018attention}, CLAM~\cite{lu2021data}, HIPT~\cite{chen2022scaling}, AttnMISL~\cite{yao2020whole}, TransMIL~\cite{shao2021transmil}, OTSurv~\cite{ren2025otsurv}, and PANTHER~\cite{song2024morphological}. For multimodal survival prediction, we compare against eight recent methods including MCAT~\cite{chen2021multimodal}, CMTA~\cite{zhou2023cross}, MOTCat~\cite{xu2023multimodal}, SurvPath~\cite{jaume2024modeling}, PIBD~\cite{zhang2024prototypical}, MMP~\cite{song2024multimodal}, MRePath~\cite{qu2025multimodal}, and LD-CVAE~\cite{zhou2025robust}.
All comparison methods were re-run with the same splits, gene tokens, and WSI features.

The results are reported in {Table~\ref{table:table1}}. As is shown, compared with unimodal methods, most multimodal approaches including ours achieve higher overall C-index, indicating complementary benefits from fusing WSIs and genomics. Among multimodal methods, TTA attains the best overall performance, outperforming the second-best by \textbf{+2.6\%} in Overall C-index. Across cohorts, TTA ranks first on BRCA, STAD, KIRC and second on BLCA and CRC. Relative to strong recent multimodal methods (e.g., MMP~\cite{song2024multimodal}, LD-CVAE~\cite{zhou2025robust}), per-dataset gains typically range from about \textbf{+0.2\% to +5.1\%}, and up to \textbf{+8.3\%} on CRC. This supports that our Together-Then-Apart design, which \emph{minimizes semantic discrepancies} during alignment and \emph{maximizes modality distinctiveness} thereafter, yields more discriminative multimodal representations for survival prediction.

\myparagraph{Analysis on the CRC Cohort.}
CRC is particularly challenging for multimodal survival modeling. 
Clinical prognosis in colorectal cancer is strongly driven by histological morphology, meaning that most predictive signals originate from WSIs. 
In this setting, strong cross-modal alignment can dilute discriminative morphological cues, causing several multimodal methods to underperform WSI-only baselines. 
TTA alleviates this issue through the \textsc{Apart} stage, which explicitly preserves WSI-specific evidence via anchor-guided contrastive regularization. 
The ablation in Table~\ref{table:table_ablation} confirms this behavior: removing \textsc{Apart} leads to a \textbf{$-$9.8\%} drop on CRC ($0.685 \rightarrow 0.587$). 
This result highlights the importance of preserving modality-specific representations in settings where predictive signals are unevenly distributed across modalities.

\begin{table*}[t]
\setlength{\arrayrulewidth}{0.1mm} 
\scriptsize
\setlength{\tabcolsep}{1.0pt}
\renewcommand{\arraystretch}{1.00}
\centering
\caption{\textbf{Ablation study on the \textsc{Together} and \textsc{Apart} stages.} C-index (mean $\pm$ std) is reported across five cohorts, with the last column showing the average across cohorts. Gray rows denote the default setting.}
\label{table:table_ablation}
\begin{tabular}{cc|cccccc}
\toprule
\textbf{TOGETHER} & \textbf{APART} & \textbf{BRCA} & \textbf{BLCA} & \textbf{STAD} & \textbf{CRC} & \textbf{KIRC} & \textbf{Overall} \\
\midrule

 & & \pmstd{0.682}{0.082} & \pmstd{0.660}{0.063} & \pmstd{0.558}{0.071} & \pmstd{0.561}{0.170} & \pmstd{0.756}{0.124} & 0.643 \\

\checkmark & & \pmstd{0.675}{0.078} & \pmstd{0.682}{0.067} & \pmstd{0.582}{0.043} & \pmstd{0.587}{0.177} & \pmstd{0.784}{0.088} & 0.662 \\

& \checkmark & \pmstd{0.683}{0.034} & \pmstd{0.651}{0.077} & \pmstd{0.585}{0.099} & \pmstd{0.625}{0.145} & \pmstd{0.785}{0.098} & 0.666 \\

\rowcolor{gray!20}
\checkmark & \checkmark & \textbf{\pmstd{0.726}{0.039}} & \textbf{\pmstd{0.662}{0.079}} & \textbf{\pmstd{0.613}{0.079}} & \textbf{\pmstd{0.685}{0.131}} & \textbf{\pmstd{0.778}{0.117}} & \textbf{0.693} \\

\bottomrule
\end{tabular}
\end{table*}

\begin{table*}[t]
\setlength{\arrayrulewidth}{0.1mm} 
\scriptsize
\setlength{\tabcolsep}{1.05pt}
\renewcommand{\arraystretch}{1.00}
\centering
\caption{\textbf{Ablation study on the \textsc{Together} stage:} shared vs. modality-specific prototypes and joint vs. modality-specific unbalanced optimal transport. C-index (mean $\pm$ std) is reported across five cohorts, with the last column showing the average across cohorts. Gray rows denote the default setting.}
% \caption{Ablation study about shared prototypes or respective prototypes, and joint unbalanced optimal transport or respective unbalanced optimal transport module in the \textsc{Together} stage.}
\label{table:table_ablation_together}
\begin{tabular}{cc|cccccc}
\toprule

\textbf{\shortstack{Shared \\ Prototypes}} & \textbf{\shortstack{joint \\ UOT}} & \textbf{BRCA} & \textbf{BLCA} & \textbf{STAD} & \textbf{CRC} & \textbf{KIRC} & \textbf{Overall} \\
\midrule

& & \pmstd{0.755}{0.025} & \pmstd{0.671}{0.025} & \pmstd{0.594}{0.063} & \pmstd{0.609}{0.199} & \pmstd{0.780}{0.108} & 0.681 \\

\checkmark & & \pmstd{0.713}{0.034} & \pmstd{0.662}{0.081} & \pmstd{0.605}{0.081} & \pmstd{0.672}{0.110} & \pmstd{0.768}{0.130} & 0.684 \\

\rowcolor{gray!20}
\checkmark & \checkmark & \textbf{\pmstd{0.726}{0.039}} & \textbf{\pmstd{0.662}{0.079}} & \textbf{\pmstd{0.613}{0.079}} & \textbf{\pmstd{0.685}{0.131}} & \textbf{\pmstd{0.778}{0.117}} & \textbf{0.693} \\

\bottomrule
\end{tabular}
\end{table*}

\begin{table*}[t]
\setlength{\arrayrulewidth}{0.1mm} 
\scriptsize
\setlength{\tabcolsep}{1.0pt}
\renewcommand{\arraystretch}{1.00}
\centering
\caption{\textbf{Ablation study on anchor refinement and contrastive regularization in the \textsc{Apart} stage.} ``Contrast. Regular.'' denotes Contrastive Regularization. C-index (mean $\pm$ std) is reported across five cohorts, with the last column showing the average across cohorts. Gray rows denote the default setting.}
% \caption{Ablation study about anchor refinement and contrastive regularization in the \textsc{Apart} stage.}
\label{table:table_ablation_apart}
\begin{tabular}{cc|cccccc}
\toprule
\textbf{\shortstack{Anchor \\ Refinement}} & \textbf{\shortstack{Contrast. \\ Regular.}} & \textbf{BRCA} & \textbf{BLCA} & \textbf{STAD} & \textbf{CRC} & \textbf{KIRC} & \textbf{Overall} \\
\midrule

& & \pmstd{0.675}{0.078} & \pmstd{0.682}{0.067} & \pmstd{0.582}{0.043} & \pmstd{0.587}{0.177} & \pmstd{0.784}{0.088} & 0.662 \\

\checkmark & & \pmstd{0.694}{0.048} & \pmstd{0.648}{0.066} & \pmstd{0.598}{0.099} & \pmstd{0.636}{0.155} & \pmstd{0.771}{0.122} & 0.669 \\

& \checkmark & \pmstd{0.688}{0.072} & \pmstd{0.678}{0.053} & \pmstd{0.589}{0.032} & \pmstd{0.627}{0.134} & \pmstd{0.778}{0.083} & 0.672 \\

\rowcolor{gray!20}
\checkmark & \checkmark & \textbf{\pmstd{0.726}{0.039}} & \textbf{\pmstd{0.662}{0.079}} & \textbf{\pmstd{0.613}{0.079}} & \textbf{\pmstd{0.685}{0.131}} & \textbf{\pmstd{0.778}{0.117}} & \textbf{0.693} \\

\bottomrule
\end{tabular}
\end{table*}

\subsection{Ablation Study}
\label{subsec:sblation}
\begin{figure}[t]
\centering
\includegraphics[width=0.9\linewidth]{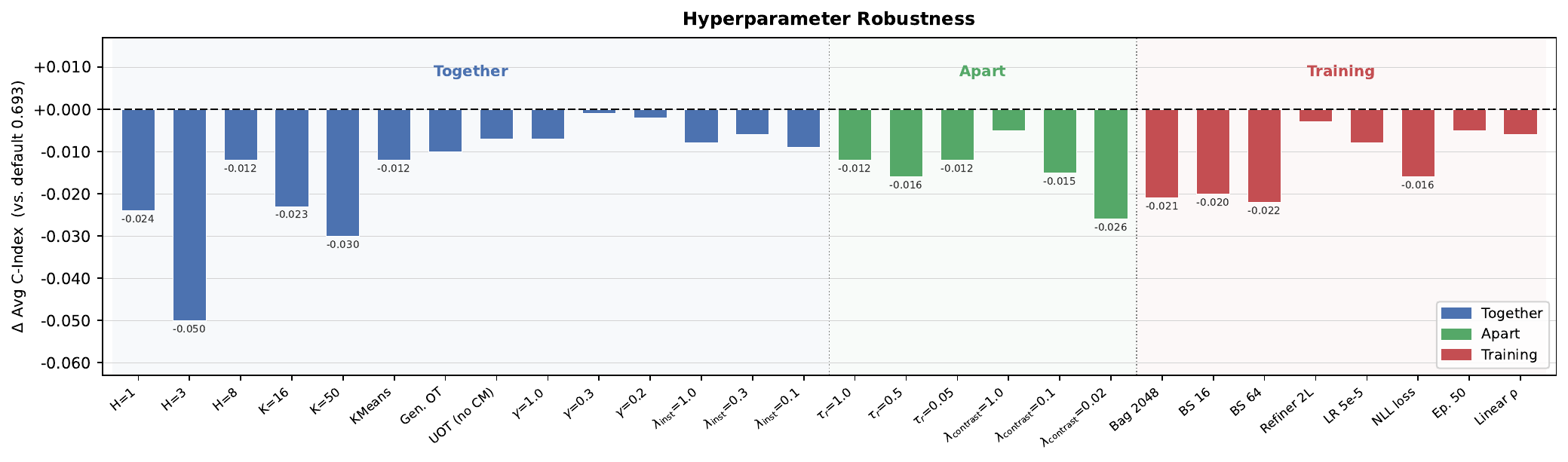}
\caption{\textbf{Hyperparameter robustness.} Each bar shows the change in average C-index across five cohorts when perturbing a single factor from the default configuration. The default Avg.\ (0.693) is indicated by the dashed line. Colors indicate groups of hyperparameters for the \textsc{Together} stage, the \textsc{Apart} stage, and training.}
% \caption{\textbf{Hyperparameter robustness.} Each bar shows the change in average C-index over five cohorts when perturbing one factor from the default configuration. The default Avg is $0.693$, shown as the dashed line. Colors group hyperparameters in \textsc{Together}, \textsc{Apart}, and training.}
\label{fig:robustness}
\end{figure}

We conduct ablation experiments to analyze the contributions of the main components in TTA.
Hyperparameter sensitivity is summarized in Fig.~\ref{fig:robustness}.
We first evaluate the individual and combined effects of the \textsc{Together} and \textsc{Apart} stages.
We then analyze key design choices within \textsc{Together}, including \emph{shared vs.\ respective prototypes}, \emph{joint vs.\ respective UOT}, and the \emph{instance-to-prototype assignment strategy} with \emph{different mass schedules}.
Finally, we study the two key components of \textsc{Apart}: \emph{anchor refinement} and \emph{contrastive regularization}.

\myparagraph{Effect of the \textsc{Together} and \textsc{Apart} Stages.}
We evaluate four variants: (i) neither stage, (ii) \textsc{Together} only, (iii) \textsc{Apart} only, and (iv) both stages.
Removing \textsc{Together} replaces shared prototypes with modality-specific prototypes and disables UOT alignment.
Removing \textsc{Apart} disables the anchor refiner and its contrastive objective.
As shown in {Table~\ref{table:table_ablation}}, enabling only \textsc{Together} improves Overall C-index by \textbf{+1.9\%}, while enabling only \textsc{Apart} yields \textbf{+2.3\%}.
Enabling both stages further increases performance to \textbf{+5.0\%}.
These results indicate that cross-modal alignment and modality-specific representation learning are complementary and jointly necessary.

\myparagraph{Shared vs.\ Respective Prototypes and Joint vs.\ Respective UOT.} 
We further analyze two design choices within the \textsc{Together} stage. When shared prototypes are disabled, each modality maintains its own prototype set. When joint UOT is disabled, optimal transport is solved independently for each modality.  With both components enabled, prototype tokens from both modalities are concatenated and matched through a joint UOT plan with a dummy sink and curriculum mass. Results in {Table~\ref{table:table_ablation_together}} show that replacing modality-specific prototypes with shared prototypes improves Overall C-index by \textbf{+0.3\%}. Further enabling joint UOT improves performance by an additional \textbf{+0.9\%}, resulting in a total gain of \textbf{+1.2\%}. This indicates that a shared prototype space combined with a unified transport plan yields more coherent cross-modal alignment.

\myparagraph{Anchor Refinement and Contrastive Regularization.}
We next examine the two components in \textsc{Apart} stage.
Table~\ref{table:table_ablation_apart} shows that anchor refinement alone improves performance by \textbf{+0.7\%}, while contrastive regularization alone yields \textbf{+1.0\%}.
Combining both components leads to a larger improvement of \textbf{+3.1\%}.
These results suggest that lightweight refinement together with contrastive coupling helps preserve modality-specific structure and prevents over-alignment.

\begin{table*}[t]
\setlength{\arrayrulewidth}{0.1mm} 
\scriptsize
\setlength{\tabcolsep}{1.0pt}
\renewcommand{\arraystretch}{1.00}
\centering
\caption{\textbf{Ablation on assignment and curriculum mass $\rho$.} C-index (mean $\pm$ std) is reported across five cohorts, with the last column showing the average across cohorts. ``curr.'' denotes curriculum. Gray rows denote the default setting.}
% \caption{\textbf{Ablations on assignment and curriculum mass.} Results are C-index (mean $\pm$ std) across five cohorts; the last column reports the average across cohorts. Gray rows indicate the default setting.}
\label{tab:robustness_key}
\begin{tabular}{c|c|ccccc|c}
\toprule
\textbf{Module} & \textbf{Setting} & \textbf{BRCA} & \textbf{BLCA} & \textbf{STAD} & \textbf{CRC} & \textbf{KIRC} & \textbf{Avg} \\
\midrule
\multirow{4}{*}{Assignment} 
& KMeans & \pmstd{0.726}{0.057} & \pmstd{0.656}{0.070} & \pmstd{0.582}{0.078} & \pmstd{0.670}{0.096} & \pmstd{0.772}{0.119} & 0.681 \\
& General OT & \pmstd{0.716}{0.036} & \pmstd{0.652}{0.085} & \pmstd{0.585}{0.082} & \pmstd{0.692}{0.112} & \pmstd{0.769}{0.124} & 0.683 \\
& {w/o curr. $\rho$} & \pmstd{0.720}{0.039} & \pmstd{0.662}{0.083} & \pmstd{0.589}{0.081} & \pmstd{0.686}{0.117} & \pmstd{0.775}{0.126} & 0.686 \\
& \cellcolor{gray!20} \textbf{{w/ curr. $\rho$}} & \cellcolor{gray!20} \pmstd{\textbf{0.726}}{0.039} & \cellcolor{gray!20} \pmstd{\textbf{0.662}}{0.079} & \cellcolor{gray!20} \pmstd{\textbf{0.613}}{0.079} & \cellcolor{gray!20} \pmstd{\textbf{0.685}}{0.131} & \cellcolor{gray!20} \pmstd{\textbf{0.778}}{0.117} & \cellcolor{gray!20} \textbf{0.693} \\
\midrule
\multirow{2}{*}{\shortstack{$\rho$ ramp-up \\ Schedule}} 
& \cellcolor{gray!20} \textbf{Sigmoid} & \cellcolor{gray!20} \pmstd{\textbf{0.726}}{0.039} & \cellcolor{gray!20} \pmstd{\textbf{0.662}}{0.079} & \cellcolor{gray!20} \pmstd{\textbf{0.613}}{0.079} & \cellcolor{gray!20} \pmstd{\textbf{0.685}}{0.131} & \cellcolor{gray!20} \pmstd{\textbf{0.778}}{0.117} & \cellcolor{gray!20} \textbf{0.693} \\
& Linear & \pmstd{0.735}{0.072} & \pmstd{0.661}{0.060} & \pmstd{0.604}{0.053} & \pmstd{0.667}{0.116} & \pmstd{0.771}{0.113} & 0.687 \\
\bottomrule
\end{tabular}
\end{table*}

\myparagraph{Instance-to-Prototype Assignment.}
We ablate the soft instance-to-prototype assignment used in \textsc{Together} (Table~\ref{tab:robustness_key}). Besides hard KMeans, we compare OT-based soft assignments (General OT and UOT): ``w/o curr.\ $\rho$'' denotes without curriculum mass $\rho$ and uses a fixed transported mass, while ``w/ curr.\ $\rho$'' ramps up $\rho$ to gradually tighten matching and route uncertain tokens to a sink early on. UOT with curriculum mass $\rho$ achieves the best Overall C-index, indicating that our model benefits from a heterogeneity-aware soft assignment with curriculum scheduling.

\myparagraph{Curriculum Mass Schedule.}
We further examine the scheduling strategy of transported mass $\rho(t)$.
Replacing the sigmoid ramp-up with a linear schedule reduces the Avg C-index from $0.693$ to $0.687$ (Table~\ref{tab:robustness_key}).
This observation suggests that gradually enforcing strict matching stabilizes training during early stages.

% \noindent\textbf{Instance-to-Prototype Assignment.} This study isolates the role of the assignment formulation in \textsc{Together}. We compare KMeans-based hard assignment, General OT, UOT without curriculum mass, and UOT with curriculum mass. As shown in Table~\ref{tab:robustness_key}, UOT with curriculum mass achieves the best Avg C-index ($0.693$), outperforming General OT ($0.683$) and KMeans-based assignment ($0.681$). These results indicate that semi-relaxed unbalanced transport together with curriculum mass yields more stable and informative instance-to-prototype assignments under multimodal heterogeneity.

% \noindent\textbf{Curriculum Mass Schedule.} We further examine the schedule used for the transported mass $\rho(t)$. Table~\ref{tab:robustness_key} shows that replacing the sigmoid ramp-up with a linear schedule reduces the Avg from $0.693$ to $0.687$. This supports the design choice of a smooth ramp-up, which delays strict matching until representations become more stable.

\myparagraph{Robustness and Hyperparameter Sensitivity.} 
We adopt a \textit{single configuration} across all datasets and folds without per-dataset tuning. Fig.~\ref{fig:robustness} shows one-at-a-time perturbations around the default configuration. Detailed results are provided in Appendix~\ref{sec:tta_addtional_experiments}. Most variants lead to only minor performance changes, indicating stable performance across a range of settings. When enabling the multi-head OT assignment, too few heads can be unstable (e.g., $H{=}3$). Meanwhile, even a single head ($H{=}1$) achieves a reasonable Avg C-index.
% We adopt a \textit{single configuration} across all datasets and folds without per-dataset tuning. Fig.~\ref{fig:robustness} shows one-at-a-time perturbations around the default configuration. Detailed results are provided in Appendix~C. Most variants lead to only minor performance changes, indicating stable performance across a range of settings. When enabling the multi-head OT assignment, too few heads can be unstable (e.g., $H{=}3$). Meanwhile, even a single head ($H{=}1$) achieves a reasonable Avg C-index.

\myparagraph{More Ablations and Visualizations.}
Additional ablation experiments and hyperparameter studies are provided in Appendix~\ref{sec:tta_addtional_experiments}. We also provide additional visualization results in {Appendix~\ref{sec:tta_additional_visualization}}.
% Additional ablation experiments and hyperparameter studies are provided in Appendix~C. We also provide additional visualization results in {Appendix~D}.

\begin{figure}[t]
\centering
\includegraphics[width=0.95\linewidth]{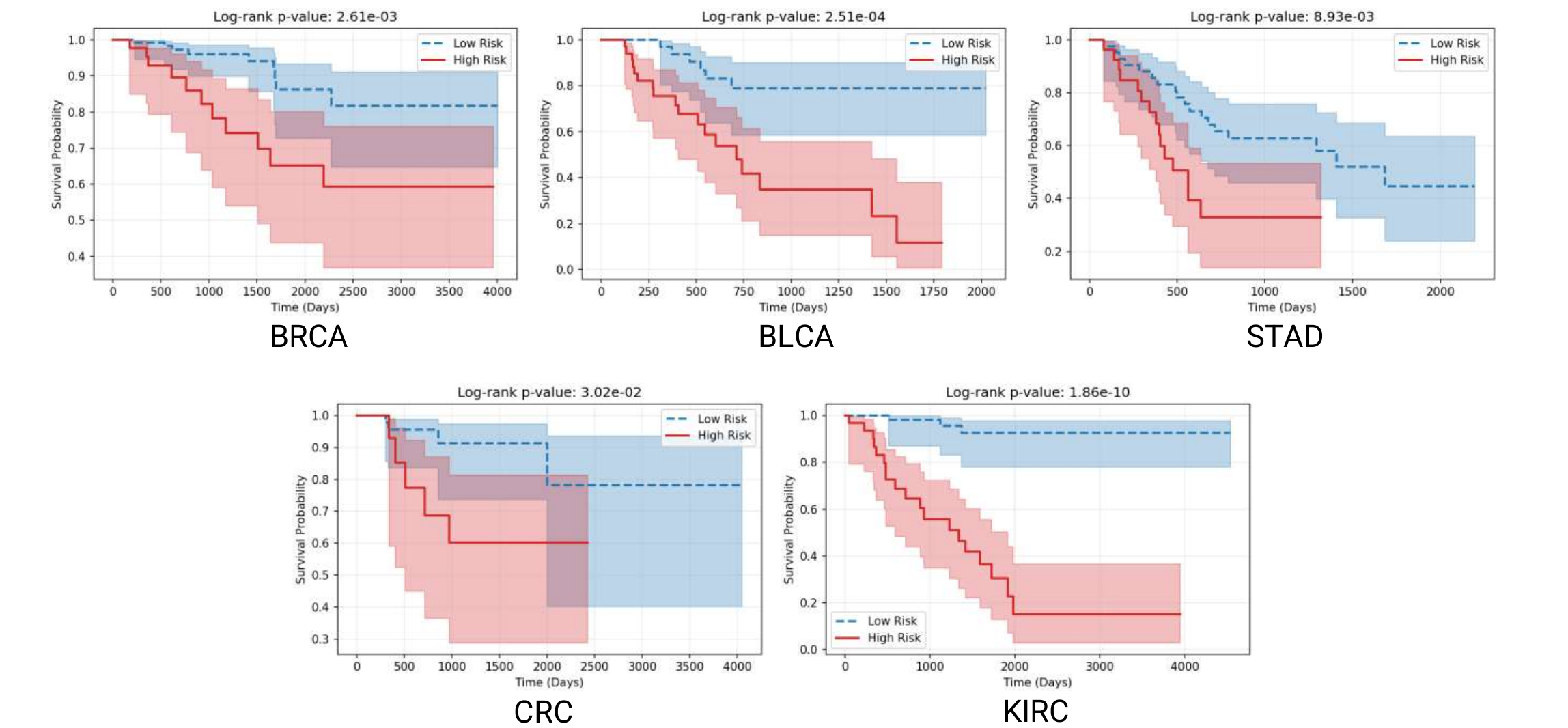}
\caption{\textbf{Kaplan-Meier curves} for the predicted high-risk (red) and low-risk (blue) groups. 
A $p$-value $<0.05$ indicates statistical significance, and shaded regions denote confidence intervals.}
\label{fig:kaplan-meier}
\end{figure}

\subsection{Stratification Visualization}
\label{sec:stratification_visual}
To further evaluate risk stratification, we perform Kaplan--Meier survival analysis based on predicted risk groups. Fig.~\ref{fig:kaplan-meier} presents Kaplan--Meier curves for all five cancer cohorts, showing clear separation between predicted high- and low-risk groups.  The log-rank test~\cite{bland2004logrank} yields statistically significant p-values across all cohorts: $2.61 \times 10^{-3}$ (BRCA), $2.51 \times 10^{-4}$ (BLCA), $8.93 \times 10^{-3}$ (STAD), $3.02 \times 10^{-2}$ (CRC), and $1.86 \times 10^{-10}$ (KIRC). All p-values are well below the $0.05$ significance threshold, confirming effective survival stratification.

\section{Conclusion}
\label{sec:conclusion}
We introduced {Together-Then-Apart (TTA)}, a framework for multimodal survival analysis that balances cross-modal alignment (\textsc{Together}) with modality-specific distinctiveness (\textsc{Apart}). 
In the \textsc{Together} stage, TTA aligns modalities through shared prototypes using an unbalanced optimal transport formulation with a curriculum mass schedule, producing stable assignments that account for intra-instance heterogeneity. 
The \textsc{Apart} stage then preserves modality-specific structure through anchor refinement and contrastive regularization, preventing representational collapse while retaining distinctive signals from each modality.
Experiments on five TCGA cohorts show that TTA consistently improves survival prediction over recent multimodal approaches and reveals interpretable cross-modal structures associated with clinical outcomes. 
More broadly, our results suggest that effective multimodal models should balance shared semantic alignment with modality-specific information, rather than forcing heterogeneous signals into a single representation. 
We hope this work encourages further study of representation learning strategies that respect both shared and modality-specific structure in multimodal biomedical data.

% \section*{Acknowledgements}
% This work was supported by . The authors thank [Name] for helpful discussions and constructive feedback on the manuscript.

\bibliographystyle{splncs04}
\bibliography{main}
\clearpage
\appendix
\setcounter{page}{1}
% Supplementary-material title block (ECCV provides \maketitlesupplementary; keep a safe fallback.)
\makeatletter
\title{\textit{Together, Then Apart}: Balancing Alignment and Distinctiveness for Multimodal Survival Analysis}
\providecommand{\maketitlesupplementary}{%
  \begin{center}
    {\Large\bfseries \@title\par}
    \vspace{0.5em}
    {\large Supplementary Material\par}
  \end{center}
}
\makeatother
\maketitlesupplementary

In this supplementary material, we provide additional theoretical analysis, experimental results, and visualizations for our work. The contents are organized as follows:
\setcounter{tocdepth}{2}
\setcounter{secnumdepth}{2}

% For safety: some templates write \authcount into *.ptc; ignore it if present.
\providecommand{\authcount}[1]{}
\startcontents[appendix]
\printcontents[appendix]{}{1}{\section*{\textbf{APPENDIX}}}

\section{Theoretical Analysis}
\label{sec:tta_theory}
\subsection{Balancing Perspective}
\label{sec:balancing_perspective}

TTA is optimized by a \emph{single} joint objective (Eq.~\eqref{eq:total_loss} in the main paper) that combines semantic alignment and modality-specific distinctiveness. This subsection provides an optimization-oriented perspective on how the \textsc{Together} and \textsc{Apart} stages interact within that joint objective.

\myparagraph{Margin interpretation of modality-specific distinctiveness.}
In the \textsc{Apart} stage, modality-specific anchors encourage the refined representation of each modality to remain closer to its own anchor than to the anchor of the other modality. Define the per-sample discrimination margin:
\begin{equation}
\Delta^m_n \;=\; s^+\!\left(\bar{h}^m_n,\,a^m\right) - s^-\!\left(\bar{h}^m_n,\,a^{\bar{m}}\right),
\end{equation}
where $s^+$ and $s^-$ are the positive and negative cosine scores defined in the main paper, and $\bar{m}$ denotes the other modality. The InfoNCE objective can be written as:
\begin{equation}
\mathcal{L}_{\text{contrast}} \;=\; \mathbb{E}\!\left[\log\!\left(1 + e^{-\Delta^m_n/\tau_r}\right)\right].
\end{equation}
Since $f(x)=\log(1+e^{-x/\tau_r})$ is strictly decreasing, minimizing $\mathcal{L}_{\text{contrast}}$ increases the expected margin $\mathbb{E}[\Delta^m_n]$. Accordingly, the contrastive term promotes modality-specific separation in the refined representation space without requiring the two modalities to be statistically independent.

\myparagraph{Remark on modality-specific information.}
The proposed framework does \emph{not} require the two modalities to carry statistically independent information. The anchor objective imposes a geometric constraint: the mean representation $\bar{h}^m_n$ of modality $m$ should satisfy $\Delta^m_n > 0$, i.e., it should lie closer to anchor $a^m$ than to $a^{\bar{m}}$. Shared biological signals, such as tumor grade reflected in both histology and gene expression, are therefore not suppressed. What is preserved is the modality-discriminative geometry of the representation space, rather than information-theoretic independence.

\myparagraph{Why the two objectives are complementary, not conflicting.}
A natural question is whether \textsc{Together} and \textsc{Apart} impose conflicting pressures on the representation. In TTA, the two stages act on \emph{different properties of the same representation}.

\textsc{Together} acts at the \emph{assignment level}: the UOT plan $Q^\star$ determines how much mass each token contributes to each shared prototype. Importantly, this does not merge the two modalities into a single vector. The two modality streams are still aggregated from their own tokens:
\begin{equation}
H^m_n=(W^m_n)^\top X^m_n,\quad m\in\{p,g\}.
\end{equation}
Thus, \textsc{Together} establishes a shared prototype coordinate system while preserving modality-specific streams.

\textsc{Apart} acts at the \emph{geometric level}. For modality $m$, the anchor objective increases the margin
\begin{equation}
\Delta^m_n=
\langle \phi(\bar h^m_n),\phi(a^m)\rangle
-\langle \phi(\bar h^m_n),\phi(a^{\bar m})\rangle .
\end{equation}
This is an ordering constraint on relative similarity: it encourages $\Delta^m_n>0$, but does not require $\bar h^m_n=a^m$. Patient-specific variation can therefore remain in the $K$ prototype tokens and in directions not fixed by this modality-margin constraint.

Finally, co-attention is applied after modality-aware refinement. It exchanges evidence across prototype tokens, but it does not impose an equality constraint such as $\|\hat H^p_n-\hat H^g_n\|^2$. Therefore, cross-modal interaction and modality-specific preservation are not contradictory: the former routes complementary evidence, while the latter regularizes the geometry of the modality streams.

\myparagraph{Gradient decomposition.}
Let $\theta_{\text{A}}$ denote parameters specific to anchor refinement (anchors, refiner $\mathcal{R}_\theta$, and the projection head), and let $\theta_{\text{T}}$ denote the remaining parameters (encoders, projections, prototypes, and the OT module). Since the instance-level loss is defined before anchor refinement, it does not depend on $\theta_{\text{A}}$. The gradients therefore decompose as:
\begin{align}
\label{eq:grad_decomp}
\nabla_{\theta_{\text{T}}}\mathcal{L}_{\text{total}}
&=\nabla_{\theta_{\text{T}}}\mathcal{L}_{\text{surv}}
+\lambda_{\text{inst}}\nabla_{\theta_{\text{T}}}\mathcal{L}_{\text{instance}}
+\lambda_{\text{contrast}}\nabla_{\theta_{\text{T}}}\mathcal{L}_{\text{contrast}}, \nonumber\\
\nabla_{\theta_{\text{A}}}\mathcal{L}_{\text{total}}
&=\nabla_{\theta_{\text{A}}}\mathcal{L}_{\text{surv}}
+\lambda_{\text{contrast}}\nabla_{\theta_{\text{A}}}\mathcal{L}_{\text{contrast}}.
\end{align}
This decomposition clarifies the respective roles of the auxiliary terms: the instance-level loss stabilizes prototype assignments in \textsc{Together}, whereas the contrastive loss shapes the anchor-based refinement in \textsc{Apart}. The survival loss couples both stages through the final prediction objective.

\myparagraph{Experimental evidence for complementarity.}
Empirically, the ablation in Table~\ref{table:table_ablation} of the main paper is consistent with this complementary interpretation: \textsc{Together} alone yields $+1.9\%$, \textsc{Apart} alone yields $+2.3\%$, and enabling both stages yields $+5.0\%$ over the base model. These observations suggest that alignment provides a stable semantic foundation for subsequent modality-discriminative refinement, while preserved modality structure in turn contributes richer features to the survival prediction objective.

\subsection{Scaling Algorithm for General OT}
\label{sec:scaling_general_ot}

The optimal transport (OT) problem can be formulated as a minimization task over transport plans. Given probability vectors $a \in \mathbb{R}^{N_{\mathrm{tot}}\times 1}$, $b \in \mathbb{R}^{K\times 1}$, along with a cost matrix $C_n \in \mathbb{R}^{N_{\mathrm{tot}}\times K}$ defined on joint space, the objective function is written as:
\begin{equation}
\label{eq:general_kantorovich_obj}
\min_{Q\in\mathbb{R}^{N_{\mathrm{tot}}\times K}_{\ge 0}} \langle Q, C_n \rangle_F + F_1(Q\mathbf{1}_K, a) + F_2(Q^\top\mathbf{1}_{N_{\mathrm{tot}}}, b)
\end{equation}
where $Q \in \mathbb{R}^{N_{\mathrm{tot}}\times K}$ denotes the transportation plan, $\langle\cdot,\cdot\rangle_F$ is the Frobenius product. $F_1$ and $F_2$ are convex marginal distribution constraints, respectively. $\mathbf{1}_K \in \mathbb{R}^{K\times 1}, \mathbf{1}_{N_{\mathrm{tot}}} \in \mathbb{R}^{N_{\mathrm{tot}}\times 1}$ are all one vectors. This is the classical Kantorovich formulation if $F_1$ and $F_2$ are equality constraints. By relaxing the marginal constraints via KL divergence or inequality penalties, the problem generalizes to the unbalanced OT as described in Section~\ref{sec:derivation_curriculum}.

To make this problem computationally tractable, Cuturi proposed entropic regularization~\cite{cuturi2013sinkhorn}. Adding the entropy term $-\epsilon H(Q)$ to objective function leads to the following formulation:
\begin{equation}
\label{eq:entropy_introduced}
\begin{aligned}
\langle Q, C_n \rangle_F - \epsilon H(Q) &= \epsilon \langle Q, C_n/\epsilon + \log Q \rangle_F \\
&= \epsilon \langle Q, \log \frac{Q}{\exp(-C_n/\epsilon)} \rangle_F \\
&= \epsilon \mathrm{KL}(Q \| \exp(-C_n/\epsilon)).
\end{aligned}
\end{equation}
Furthermore, Eq.~\eqref{eq:entropy_introduced} can be reformulated as:
\begin{equation}
\label{eq:entropic_ot_obj}
\begin{aligned}
\min_{Q\in\mathbb{R}^{N_{\mathrm{tot}}\times K}_{\ge 0}} & \epsilon \mathrm{KL}(Q \| \exp(-C_n/\epsilon))\\ & + F_1(Q\mathbf{1}_K, a) + F_2(Q^\top\mathbf{1}_{N_{\mathrm{tot}}}, b).
\end{aligned}
\end{equation}
Define the proximal operator as:
\begin{equation}
\label{eq:proximal_op_def}
\mathrm{prox}_{f/\epsilon}^{KL}(y; z) = \arg\min_{x\ge 0} f(x, z) + \epsilon \mathrm{KL}(x \| y)
\end{equation}
where $z$ is the fixed parameter of the function $f$. In our case, $z$ corresponds to the marginal distributions while $f$ represents the associated marginal constraints $F_1$ or $F_2$. Then Eq.~\eqref{eq:entropic_ot_obj} can be solved approximately using Alg.~\ref{alg:generalized_scaling}.

\begin{algorithm}[t]
\caption{Generalized Scaling Algorithm}
\label{alg:generalized_scaling}
\begin{algorithmic}[1]
\State \textbf{Input:} Cost $C_n$, regularization $\epsilon > 0$, marginals $a \in \mathbb{R}^{N_{\mathrm{tot}}}_{\ge 0}$, $b \in \mathbb{R}^{K}_{\ge 0}$
\State $G \leftarrow \exp(-C_n/\epsilon) \quad \triangleright G_{ij} = e^{-C_{n,ij}/\epsilon}$
\State $v \leftarrow \mathbf{1}_{K}$
\While{not converged}
    \State $x \leftarrow G v$
    \State $\tilde u \leftarrow \mathrm{prox}_{F_1/\epsilon}^{KL}(x; a)$
    \State $u \leftarrow \tilde u \oslash x \quad \triangleright \text{elementwise division}$
    \State $y \leftarrow G^\top u$
    \State $\tilde v \leftarrow \mathrm{prox}_{F_2/\epsilon}^{KL}(y; b)$
    \State $v \leftarrow \tilde v \oslash y$
\EndWhile
\State \textbf{Return:} $Q^\star = \mathrm{diag}(u) G \mathrm{diag}(v)$
\end{algorithmic}
\end{algorithm}
These updates can be interpreted as Bregman projections with respect to the KL divergence onto convex sets defined by the marginal constraints. Alternating such projections is guaranteed to converge, and the diagonal scaling form makes each iteration linear in the number of nonzero entries of $G$. The entropic regularization enforces strict positivity, prevents sparsity and collapse of the transport plan, and enhances numerical stability. Intuitively, the scaling vectors $u, v$ can be viewed as per-row and per-column adjustment factors, respectively. Multiplying by $u$ rescales entire rows to match $a$, while multiplying by $v$ rescales columns to align with $b$. The alternating iterations drive the transport plan $Q$ toward satisfying the marginal structure.

As a result, whenever an optimal transport problem can be reformulated with suitable marginal constraints into the form of Eq.~\eqref{eq:general_kantorovich_obj}, the corresponding proximal operators can be derived as in Eq.~\eqref{eq:proximal_op_def}. This allows the problem to be efficiently solved using Alg.~\ref{alg:generalized_scaling}.

\subsection{From Standard OT to UOT with Curriculum Mass}
\label{sec:derivation_curriculum}

When strict equality constraints are not enforced, one may allow mass to be created or discarded. This leads to the unbalanced OT formulation where deviations from the marginals are penalized by a KL divergence. Assuming a uniform source distribution, Eq.~\eqref{eq:general_kantorovich_obj} can be expressed as:
\begin{equation}
\begin{aligned}
\min_{Q\in\Pi} \quad & \langle Q, C_n \rangle_F + \gamma \mathrm{KL}\left(Q^\top \mathbf{1}_{N_{\mathrm{tot}}} \,\Big\|\, \frac{1}{K}\mathbf{1}_K\right) \\
\text{s.t.} \quad & \Pi = \left\{ Q \in \mathbb{R}^{N_{\mathrm{tot}}\times K}_{\ge 0} \mid Q \mathbf{1}_K = \frac{1}{N_{\mathrm{tot}}}\mathbf{1}_{N_{\mathrm{tot}}} \right\},
\end{aligned}
\end{equation}
where $\gamma$ is the regularization weight factor. Here, the row sums are fixed to the uniform source distribution $\frac{1}{N_{\mathrm{tot}}}\mathbf{1}_{N_{\mathrm{tot}}}$, while the column sums are softly penalized toward the uniform target distribution $\frac{1}{K}\mathbf{1}_K$.

Although unbalanced OT relaxes the marginal constraints, it still penalizes discrepancies between the transported and target mass. As a result, especially early in training, even ambiguous or noisy features are still encouraged to be moved, potentially degrading the quality of the solution. To address this limitation, we adopt the UOT with curriculum mass formulation, which explicitly controls the amount of total transported mass. Instead of hard-thresholding unreliable features, UOT with curriculum mass allows the model to reweigh and selectively transport a subset of the source samples by solving:
\begin{equation}
\label{eq:partial_ot_definition}
\begin{aligned}
\min_{Q\in\Pi} \quad & \langle Q, C_n \rangle_F + \gamma \mathrm{KL}\left(Q^\top \mathbf{1}_{N_{\mathrm{tot}}} \,\Big\|\, \frac{\rho}{K}\mathbf{1}_K\right) \\
\text{s.t.} \quad & \Pi = \bigg\{ Q \in \mathbb{R}^{N_{\mathrm{tot}}\times K}_{\ge 0} \;\bigg|\; Q \mathbf{1}_K \le \frac{1}{N_{\mathrm{tot}}}\mathbf{1}_{N_{\mathrm{tot}}}, \mathbf{1}_{N_{\mathrm{tot}}}^\top Q \mathbf{1}_K = \rho \bigg\},
\end{aligned}
\end{equation}
where $N_{\mathrm{tot}}$ is the uniform source feature count and $K$ is the number of target prototypes. $\rho$ specifies the total transported mass and will increase gradually. Intuitively, UOT with curriculum mass still respects the distributional structure but enables progressive selection of reliable samples. Low-cost correspondences are favored first, while noisier or ambiguous features can be safely ignored or deferred until $\rho$ increases. This mechanism provides a principled way to suppress noise while guiding the optimization toward a globally consistent transport plan.

\subsection{Reformulation and Proof of Equivalence}
\label{sec:proof_equivalence}

To solve UOT with curriculum mass efficiently using scaling algorithms, we follow prior work~\cite{chapel2020partial,ICLR2024_3d037913} and reformulate the problem. The key idea is to introduce a slack column into the marginal distribution to absorb the unselected mass $1 - \rho$, thereby turning the global mass constraint into a marginal one. Specifically, the slack column is denoted as $\eta \in \mathbb{R}^{N_{\mathrm{tot}}\times 1}$ to absorb the remaining mass and form the extended coupling:
\begin{equation}
\tilde Q = [Q, \eta] \in \mathbb{R}^{N_{\mathrm{tot}}\times (K+1)}, \quad \tilde C_n = [C_n, \mathbf{0}_{N_{\mathrm{tot}}}].
\end{equation}
Imposing row-sum equality to the uniform source $a = \frac{1}{N_{\mathrm{tot}}}\mathbf{1}_{N_{\mathrm{tot}}}$ and total-mass accounting, we get:
\begin{equation}
\tilde Q \mathbf{1}_{K+1} = \frac{1}{N_{\mathrm{tot}}}\mathbf{1}_{N_{\mathrm{tot}}}, \quad \mathbf{1}_{N_{\mathrm{tot}}}^\top \eta = 1 - \rho, \quad \mathbf{1}_{N_{\mathrm{tot}}}^\top Q \mathbf{1}_K = \rho,
\end{equation}
Thus,
\begin{equation}
\tilde Q^\top \mathbf{1}_{N_{\mathrm{tot}}} = \begin{bmatrix} Q^\top \mathbf{1}_{N_{\mathrm{tot}}} \\ \eta^\top \mathbf{1}_{N_{\mathrm{tot}}} \end{bmatrix} = \begin{bmatrix} Q^\top \mathbf{1}_{N_{\mathrm{tot}}} \\ 1 - \rho \end{bmatrix}.
\end{equation}
Let the target column-mass prior be:
\begin{equation}
\tilde b(\rho) = \begin{bmatrix} \frac{\rho}{K}\mathbf{1}_K \\ 1-\rho \end{bmatrix} \in \mathbb{R}^{K+1},
\end{equation}
Then, the KL-penalized unbalanced surrogate of UOT with curriculum mass is:
\begin{equation}
\label{eq:surrogate_uot}
\begin{aligned}
\min_{\tilde Q \in \Phi} \quad & \langle \tilde Q, \tilde C_n \rangle_F + \gamma \mathrm{KL}(\tilde Q^\top \mathbf{1}_{N_{\mathrm{tot}}} \| \tilde b(\rho)) \\
\text{s.t.} \quad & \Phi = \{ \tilde Q \in \mathbb{R}^{N_{\mathrm{tot}}\times (K+1)}_{\ge 0} \mid \tilde Q \mathbf{1}_{K+1} = \frac{1}{N_{\mathrm{tot}}}\mathbf{1}_{N_{\mathrm{tot}}} \}.
\end{aligned}
\end{equation}
However, the KL term is soft. Eq.~\eqref{eq:surrogate_uot} does not guarantee the mass of the last column to be strictly $1 - \rho$. To recover the exact constraint of UOT with curriculum mass, a weighted $KL$ constraint is employed to control the constraint strength for each class:
\begin{equation}
\hat{KL}(\tilde Q^\top \mathbf{1}_{N_{\mathrm{tot}}} \| \tilde b(\rho); \hat{\gamma}) = \sum_{i=1}^{K+1} \hat{\gamma}_i [\tilde Q^\top \mathbf{1}_{N_{\mathrm{tot}}}]_i \log \frac{[\tilde Q^\top \mathbf{1}_{N_{\mathrm{tot}}}]_i}{[\tilde b(\rho)]_i},
\end{equation}
with,
\begin{equation}
\hat{\gamma} = \begin{bmatrix} \gamma \mathbf{1}_K \\ +\infty \end{bmatrix}.
\end{equation}
This yields the final equivalent formulation:
\begin{equation}
\label{eq:final_equivalent_formulation}
\begin{aligned}
\min_{\tilde Q \in \Phi} \quad & \langle \tilde Q, \tilde C_n \rangle_F + \hat{KL}(\tilde Q^\top \mathbf{1}_{N_{\mathrm{tot}}} \| \tilde b(\rho); \hat{\gamma}) \\
\text{s.t.} \quad & \Phi = \{ \tilde Q \in \mathbb{R}^{N_{\mathrm{tot}}\times (K+1)}_{\ge 0} \mid \tilde Q \mathbf{1}_{K+1} = \frac{1}{N_{\mathrm{tot}}}\mathbf{1}_{N_{\mathrm{tot}}} \}
\end{aligned}
\end{equation}
The weighted KL makes the slack mass non-negotiable while keeping the real columns softly regularized. So low-cost correspondences are selected first, and ambiguous features can be safely left in the slack. The extended optimal plan is consistent with the original one, and the first $K$ columns of the extended solution align with the optimal plan of the UOT with curriculum mass problem. The proof is provided below.

\myparagraph{Proof of Equivalence with UOT with Curriculum Mass.}
In this section, we present the full proof that $\hat Q^\star$, corresponding to the first $K$ columns of the extended optimal transport plan $\tilde Q^\star$, coincides with the optimal plan $Q^\star$ of the UOT with curriculum mass problem.

\noindent\textit{Proof.} Assume the optimal extended plan is:
\begin{equation}
\tilde Q^\star = [\hat Q^\star, \eta^\star] \in \mathbb{R}^{N_{\mathrm{tot}}\times (K+1)}, \quad \hat Q^\star \in \mathbb{R}^{N_{\mathrm{tot}}\times K}.
\end{equation}
The weighted KL penalty expands as:
\begin{equation}
\begin{aligned}
\hat{KL}(\tilde Q^{\star\top} \mathbf{1}_{N_{\mathrm{tot}}} \| \tilde b(\rho);& \hat{\gamma})\\&=\sum_{i=1}^{K} \gamma_i [\hat Q^{\star\top} \mathbf{1}_{N_{\mathrm{tot}}}]_i \log \frac{[\hat Q^{\star\top} \mathbf{1}_{N_{\mathrm{tot}}}]_i}{[\frac{\rho}{K}\mathbf{1}_K]_{i}} \\&+ \gamma_{K+1} \eta^{\star\top} \mathbf{1}_{N_{\mathrm{tot}}} \log \frac{\eta^{\star\top} \mathbf{1}_{N_{\mathrm{tot}}}}{1 - \rho} \\
&= \gamma \mathrm{KL}(\hat Q^{\star\top} \mathbf{1}_{N_{\mathrm{tot}}} \| \frac{\rho}{K}\mathbf{1}_K) \\&+ \gamma_{K+1} \eta^{\star\top} \mathbf{1}_{N_{\mathrm{tot}}} \log \frac{\eta^{\star\top} \mathbf{1}_{N_{\mathrm{tot}}}}{1 - \rho}.
\end{aligned}
\end{equation}
Taking the limit $\gamma_{K+1} \to +\infty$ forces the slack column to satisfy $\eta^{\star\top} \mathbf{1}_{N_{\mathrm{tot}}} = 1 - \rho$, otherwise the objective would diverge.
By construction, the extended plan satisfies the row constraint:
\begin{equation}
\tilde Q^\star \mathbf{1}_{K+1} = \frac{1}{N_{\mathrm{tot}}}\mathbf{1}_{N_{\mathrm{tot}}}.
\end{equation}
This can be written as:
\begin{equation}
{\hat Q^\star} \mathbf{1}_K + \eta^\star = \frac{1}{N_{\mathrm{tot}}}\mathbf{1}_{N_{\mathrm{tot}}}, \quad \eta^\star \ge 0,
\end{equation}
which implies:
\begin{equation}
{\hat Q^\star} \mathbf{1}_K \le \frac{1}{N_{\mathrm{tot}}}\mathbf{1}_{N_{\mathrm{tot}}}.
\end{equation}
In addition, the total transported mass of the first $K$ columns is:
\begin{equation}
\mathbf{1}_{N_{\mathrm{tot}}}^\top {\hat Q^\star} \mathbf{1}_K = \mathbf{1}_{N_{\mathrm{tot}}}^\top \tilde Q^\star \mathbf{1}_{K+1} - \mathbf{1}_{N_{\mathrm{tot}}}^\top \eta^\star = 1 - (1 - \rho) = \rho.
\end{equation}
Therefore,
\begin{equation}
{\hat Q^\star} \in \{ Q \in \mathbb{R}^{N_{\mathrm{tot}}\times K} \mid Q \mathbf{1}_K \le \frac{1}{N_{\mathrm{tot}}}\mathbf{1}_{N_{\mathrm{tot}}}, \mathbf{1}_{N_{\mathrm{tot}}}^\top Q \mathbf{1}_K = \rho \},
\end{equation}
which is precisely the feasible set of the UOT with curriculum mass problem.
Lastly, the cost of the extended problem is:
\begin{equation}
\label{eq:extended_ot}
\begin{aligned}
& \langle \tilde Q^\star, \tilde C_n \rangle_F + \hat{KL}(\tilde Q^{\star\top} \mathbf{1}_{N_{\mathrm{tot}}} \| \tilde b(\rho); \hat{\gamma}) \\&= \langle [{\hat Q^\star}, \eta^\star], [C_n, \mathbf{0}_{N_{\mathrm{tot}}}] \rangle_F + \gamma \mathrm{KL}(\hat Q^{\star\top} \mathbf{1}_{N_{\mathrm{tot}}} \| \frac{\rho}{K}\mathbf{1}_K) \\
&= \langle {\hat Q^\star}, C_n \rangle_F + \gamma \mathrm{KL}(\hat Q^{\star\top}\mathbf{1}_{N_{\mathrm{tot}}} \| \frac{\rho}{K}\mathbf{1}_K)
\end{aligned}
\end{equation}
This is exactly the objective of the UOT with curriculum mass problem as in Eq.~\eqref{eq:partial_ot_definition} evaluated at ${\hat Q^\star}$.
If ${\hat Q^\star}$ achieves a lower cost than $Q^\star$ for the initial UOT with curriculum mass formula, it contradicts the optimality of $Q^\star$ (as $Q^\star$ is defined as the optimal solution).
Conversely, if $Q^\star$ had strictly lower cost for Eq.~\eqref{eq:extended_ot}, then $\hat Q^{\star}$ would no longer achieve the optimum, which would contradict the optimality of $\tilde Q^\star$.
As a result, by convexity of the objective, ${\hat Q^\star} = Q^\star$. Dropping the last column of $\tilde Q^\star$, we achieve the optimal transport plan for the UOT with curriculum mass problem. 

\subsection{Solver for UOT}
\label{sec:solver_uot}

Adding an entropy regularization term $-\epsilon H(\tilde Q)$ to Eq.~\eqref{eq:final_equivalent_formulation} also enables an efficient scaling algorithm. We denote:
\begin{equation}
G = \exp(-\tilde C_n / \epsilon), \quad f = \frac{\hat \gamma}{\hat \gamma + \epsilon}, \quad \alpha = \frac{1}{N_{\mathrm{tot}}}\mathbf{1}_{N_{\mathrm{tot}}}.
\end{equation}
The optimal plan admits the standard scaling form:
\begin{equation}
\tilde Q^\star = \mathrm{diag}(u) G \mathrm{diag}(v).
\end{equation}
\textit{Proof.} As in Section~\ref{sec:scaling_general_ot}, the main step is to compute the proximal operators corresponding to the constraints. To this end, let us first restate Eq.~\eqref{eq:final_equivalent_formulation} in a more general form:
\begin{equation}
\begin{aligned}
\min_{\tilde Q \in \Phi} \quad & \epsilon \mathrm{KL}(\tilde Q \| \exp(- \tilde C_n / \epsilon)) + \hat{KL}(\tilde Q^\top \mathbf{1}_{N_{\mathrm{tot}}} \| \tilde b(\rho); \hat \gamma), \\
\text{s.t.} \quad & \Phi = \{ \tilde Q \in \mathbb{R}^{N_{\mathrm{tot}}\times (K+1)}_{\ge 0} \mid \tilde Q \mathbf{1}_{K+1} = \alpha \}
\end{aligned}
\end{equation}
where $\tilde C_n$ is the cost matrix, $\alpha$ is the source marginal.
The equality constraint $\tilde Q \mathbf{1}_{K+1} = \alpha$ can be expressed as the indicator:
\begin{equation}
F_1(x; \alpha) = \begin{cases} 0, & x = \alpha, \\ +\infty, & \text{otherwise}. \end{cases}
\end{equation}
This directly gives $\mathrm{prox}_{F_1/\epsilon}^{KL}(y; \alpha) = \alpha$.
For the weighted KL penalty, the proximal operator is defined as:
\begin{equation}
\begin{aligned}
\mathrm{prox}_{F_2/\epsilon}^{KL}(y; \tilde b(\rho)) &= \arg\min_{x \ge 0} \hat{KL}(x \| \tilde b(\rho); \hat \gamma) + \epsilon \mathrm{KL}(x \| y) \\
= \arg&\min_{x \ge 0} \sum_{i=1}^{K+1} \hat \gamma_i (x_i \log \frac{x_i}{[\tilde b(\rho)]_i} - x_i + [\tilde b(\rho)]_i) \\&+ \epsilon (x_i \log \frac{x_i}{y_i} - x_i + y_i).
\end{aligned}
\end{equation}
After dropping constants independent of $x$ and regrouping terms, we obtain:
\begin{equation}
\begin{aligned}
\mathrm{prox}_{F_2/\epsilon}^{KL}(y; \tilde b(\rho)) = \arg\min_{x \ge 0} \sum_{i=1}^{K+1} (\hat \gamma_i + \epsilon) x_i \log x_i \\- (\hat \gamma_i \log [\tilde b(\rho)]_i + \hat \gamma_i + \epsilon \log y_i + \epsilon) x_i.
\end{aligned}
\end{equation}
Consider the generic function $g(z) = c_1 z \log z - c_2 z$ with $c_1 > 0$. Its derivative is $g'(z) = c_1(1 + \log z) - c_2$, hence the minimizer is $z^\star = \exp(\frac{c_2-c_1}{c_1})$. Applying this result gives:
\begin{equation}
\begin{aligned}
x_i^\star &= \exp\left(\frac{\hat \gamma_i \log [\tilde b(\rho)]_i + \epsilon \log y_i}{\hat \gamma_i + \epsilon}\right) \\
&= [\tilde b(\rho)]_i^{\frac{\hat \gamma_i}{\hat \gamma_i + \epsilon}} y_i^{\frac{\epsilon}{\hat \gamma_i + \epsilon}}.
\end{aligned}
\end{equation}
In vector notation, we write:
\begin{equation}
x = \tilde b(\rho)^{\circ f}  y^{\circ (1-f)}, \quad f = \frac{\hat \gamma}{\hat \gamma + \epsilon},
\end{equation}
where $\circ$ denotes the element-wise power.
Now, substituting the two proximal operators into the general scaling algorithm yields the updates:
\begin{equation}
u \leftarrow \frac{\alpha}{G v}, \quad v \leftarrow \left( \frac{\tilde b(\rho)}{G^\top u} \right)^{\circ f},
\end{equation}
where $G = \exp(-\tilde C_n / \epsilon)$. 
The pseudo-code of the scaling algorithm for UOT with curriculum mass is provided in Algorithm~\ref{alg:scaling_final}.

\begin{algorithm}[t]
\caption{Scaling Algorithm for UOT with Curriculum Mass}
\label{alg:scaling_final}
\begin{algorithmic}[1]
\State \textbf{Input:} Cost matrix $C_n$, regularization $\epsilon$, KL weight $\gamma$, curriculum mass $\rho$, $N_{\mathrm{tot}}$, $K$, a large value $\iota$.
\State \textbf{Initialize:}
\State \quad $\tilde C_n \leftarrow [C_n, \mathbf{0}_{N_{\mathrm{tot}}}]$
\State \quad $\hat \gamma \leftarrow [\gamma, \dots, \gamma, \iota]^\top$
\State \quad $\tilde b \leftarrow [\frac{\rho}{K}\mathbf{1}_K; 1-\rho]^\top$ \Comment{Target Marginal}
\State \quad $a \leftarrow \frac{1}{N_{\mathrm{tot}}}\mathbf{1}_{N_{\mathrm{tot}}}$ \Comment{Source Marginal $\alpha$}
\State \quad $v \leftarrow \mathbf{1}_{K+1}$ \Comment{Col Scaling Vector}
\State \quad $G \leftarrow \exp(-\tilde C_n / \epsilon)$
\State \quad $f \leftarrow \frac{\hat \gamma}{\hat \gamma + \epsilon}$
\While{$v$ does not converge}
    \State $u \leftarrow \frac{a}{G v}$ \Comment{Row Update (Exact Constraint)}
    \State $v \leftarrow (\frac{\tilde b}{G^\top u})^{\circ f}$ \Comment{Col Update (Relaxed Constraint)}
    \State \Comment{Note: Slack col has $f \approx 1$ due to $\iota \to \infty$, enforcing hard constraint.}
\EndWhile
\State $\tilde Q \leftarrow \mathrm{diag}(u) G \mathrm{diag}(v)$
\State \textbf{Return:} $\tilde Q[{:, :K}]$
\end{algorithmic}
\end{algorithm}

\subsection{Loss Function for Survival Analysis}
\label{sec:loss_survival}
We introduce the two survival loss functions used in this work.

\myparagraph{Discrete-time Negative Log-likelihood (NLL).}
For sample $n$ with discrete time index $y_n$ and event indicator $\delta_n\in\{0,1\}$ ($\delta_n=1$ if the event occurs, $0$ if censored), let $h_{n,t}\in(0,1)$ denote the per-interval hazard and:
\[
S_{n,t}=\prod_{j=1}^{t}\big(1-h_{n,j}\big),\qquad S_{n,0}=1
\]
be the discrete survival function. The per-sample NLL is:
\begin{equation}
\begin{aligned}
\label{eq:disc_surv_nll}
\ell^{\mathrm{NLL}}_n
= -\,\delta_n\big(\log S_{n,y_n}+\log h_{n,y_n}\big)\;\\-\;(1-\delta_n)\,\log S_{n,y_n+1},
\end{aligned}
\end{equation}
and the batch loss is $\mathcal{L}_{\mathrm{NLL}}=\frac{1}{N}\sum_{n=1}^{N}\ell^{\mathrm{NLL}}_n$.

\myparagraph{Cox Partial Likelihood.}
Let $r_n\in\mathbb{R}$ be the predicted log-risk for sample $n$, and define the risk set $\mathcal{R}_i=\{\,j:\; y_j\ge y_i\,\}$. The negative average Cox partial log-likelihood is:
\begin{equation}
\label{eq:cox_partial_loss}
\mathcal{L}_{\mathrm{Cox}}
= -\,\frac{1}{\sum_{i=1}^{N}\delta_i}\sum_{i:\,\delta_i=1}
\Bigg[\, r_i \;-\; \log\!\!\sum_{j\in\mathcal{R}_i}\exp(r_j)\Bigg].
\end{equation}

\section{TTA Implementation Details}
\label{sec:tta_imple_details}

\begin{table}[t]
\setlength{\arrayrulewidth}{0.1mm}

\setlength{\tabcolsep}{1.0pt}
\renewcommand{\arraystretch}{1.00}
\centering
\caption{Experimental configurations. }
\label{table:impl_details}
\begin{tabular}{l l}
\toprule
\textbf{Item } & \textbf{Value} \\
\midrule
Patch size, magnification & $256{\times}256$, $20{\times}$ \\
Sampled patches per slide (bag size) & 4096 \\
Batch size & 32 \\
Max epochs & 30 \\
Optimizer, learning rate, weight decay & AdamW, $1{\times}10^{-4}$, $1{\times}10^{-5}$ \\
LR scheduler & cosine \\
Dropout & $0.3$ \\
Clip norm & $5.0$ \\
\midrule
Shared prototypes $K$ & $32$ \\
Shared space dimension $D'$ & $256$ \\
Prototype logits temperature $\tau_{\mathrm{shared}}$ & $0.5$ \\
Target KL weight $\gamma$ & $0.1$ \\
SK multi‑head numbers & $5$ \\
Instance loss weight $\lambda_{\mathrm{inst}}$ & $0.5$ \\
Softmax-OT mixing coefficient $\beta_{\mathrm{mix}}$ & $0.5$ \\
\midrule
Contrastive loss weight $\lambda_{\mathrm{contrast}}$ & $0.5$ \\
Modality weights $\lambda_{\mathrm{wsi}}, \lambda_{\mathrm{gen}}$ & $1, 1$ \\
Contrast temperature $\tau_r$ & $0.1$ \\
Refiner layers & $1$ \\
Co‑attention layers & $1$ \\
\bottomrule
\end{tabular}
\end{table}

Following Sec.~\ref{sec:tta_theory}, we compute the UOT pseudo-label plan on the concatenated pathology and genomics tokens after augmenting the cost matrix with a zero-cost sink column. The curriculum mass $\rho(t)$ follows Eq.~\eqref{eq:rho}, where $t$ is the current iteration and $T$ is the ramp-up horizon, and gradually increases as training proceeds. After solving the extended problem, we remove the sink column and split the remaining transport matrix into WSI and genomics blocks, which are used as soft pseudo-labels for the two modalities. For batched training, each WSI is represented by a fixed bag of 4096 tokens: shorter bags are zero-padded, longer bags are uniformly subsampled, and an attention mask prevents padded tokens from contributing to prototype logits or aggregation. In the shared-prototype setting, all heads share a single learnable prototype bank, while each head uses its own modality-specific projection into the shared space. For prototype aggregation, the OT-derived pseudo-labels are mixed with the softmax weights using $\beta_{\mathrm{mix}}$ and then row-normalized before pooling. The instance-level soft cross-entropy is computed separately for WSI and genomics against these OT-derived targets and then combined with weights $\lambda_{\mathrm{wsi}}$ and $\lambda_{\mathrm{gen}}$. Other configurations are summarized in Table~\ref{table:impl_details}.

\section{Additional Experiments and Analysis}
\label{sec:tta_addtional_experiments}
\subsection{Ablations in \textsc{Together} stage}
\begin{figure}[ht]
    \centering
    \includegraphics[width=0.98\linewidth]{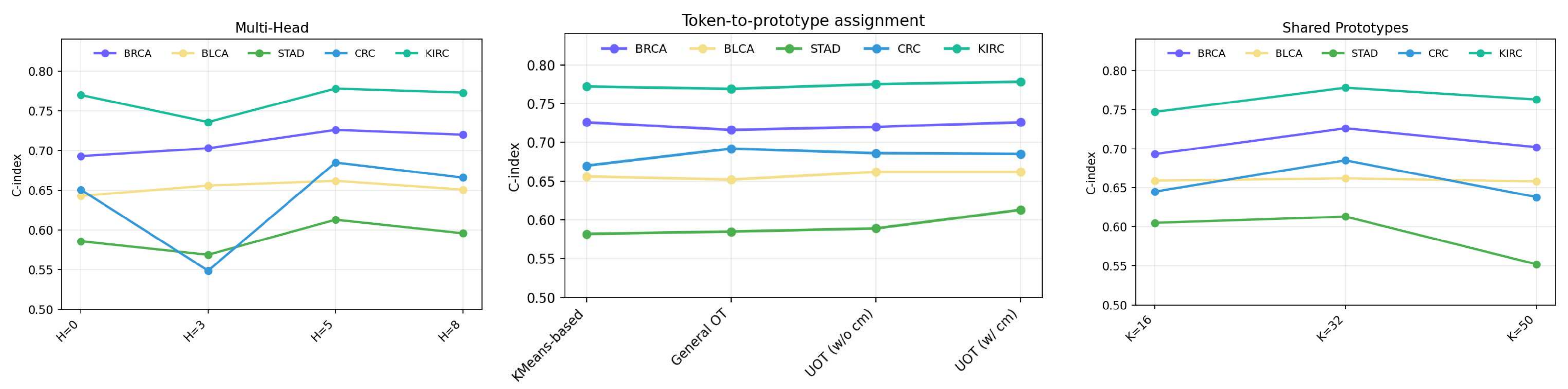}
    \caption{Ablations and hyperparameter experiments in the \textsc{Together} stage: multi-head, instance-to-prototype assignment, number of shared prototypes.}
    \label{fig:together ablations_1}
\end{figure}

\begin{figure}[t]
    \centering
    \includegraphics[width=0.9\linewidth]{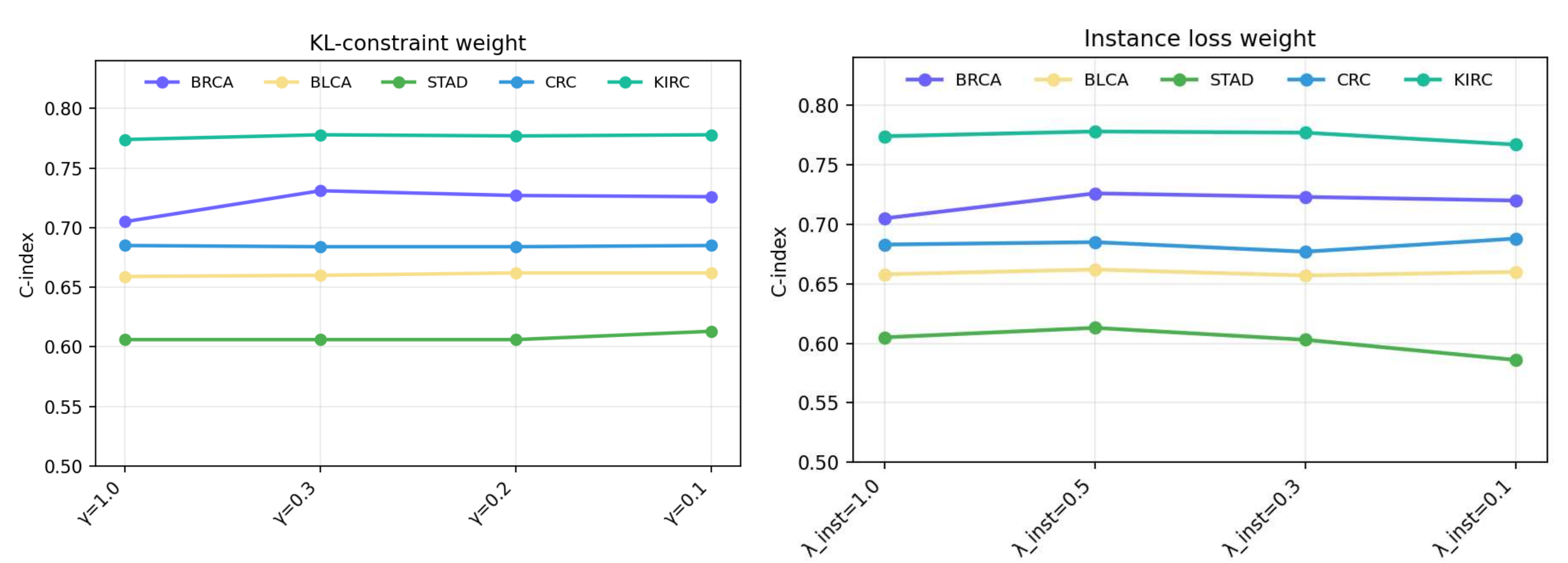}
    \caption{Hyperparameter experiments in the \textsc{Together} stage: KL-constraint weight and instance loss weight.}
    \label{fig:together ablations_2}
\end{figure}

We provide more detailed cohort-wise ablations and hyperparameter analysis for the \textsc{Together} stage. Here we focus on dataset-level behavior and hyperparameter effects beyond the summary ablations in the main paper. The results are shown in Table~\ref{tab:together_ablation_detailed_1}, Fig~\ref{fig:together ablations_1}, and Fig~\ref{fig:together ablations_2}.
\begin{table*}[t]
\setlength{\arrayrulewidth}{0.1mm} 
\scriptsize
\setlength{\tabcolsep}{1.0pt}
\renewcommand{\arraystretch}{1.00}
\centering
\caption{Ablations and hyperparameter experiments in the \textsc{Together} stage: multi-head, instance-to-prototype assignment method, number of shared prototypes, KL-constraint weight, and instance loss weight. Results are C-index (mean $\pm$ std).}
\label{tab:together_ablation_detailed_1}
\begin{tabular}{c|c|ccccc|c}
\toprule
\textbf{Module} & \textbf{Settings} & \textbf{BRCA} & \textbf{BLCA} & \textbf{STAD} & \textbf{CRC} & \textbf{KIRC} & \textbf{Avg} \\
\midrule
\multirow{4}{*}{\cellcolor{white}Multi-Head} 
& \makecell{H=1 (Single)} & \pmstd{0.693}{0.066} & \pmstd{0.643}{0.073} & \pmstd{0.586}{0.057} & \pmstd{0.651}{0.108} & \pmstd{0.770}{0.106} & 0.669 \\
& H=3    & \pmstd{0.703}{0.066} & \pmstd{0.656}{0.065} & \pmstd{0.569}{0.065} & \pmstd{0.549}{0.165} & \pmstd{0.736}{0.103} & 0.643 \\
& \cellcolor{gray!20} H=5    & \cellcolor{gray!20} \pmstd{\textbf{0.726}}{0.039} & \cellcolor{gray!20} \pmstd{\textbf{0.662}}{0.079} & \cellcolor{gray!20} \pmstd{\textbf{0.613}}{0.079} & \cellcolor{gray!20} \pmstd{\textbf{0.685}}{0.131} & \cellcolor{gray!20} \pmstd{\textbf{0.778}}{0.117} & \cellcolor{gray!20} \textbf{0.693} \\
& H=8    & \pmstd{0.720}{0.064} & \pmstd{0.651}{0.071} & \pmstd{0.596}{0.054} & \pmstd{0.666}{0.136} & \pmstd{0.773}{0.134} & 0.681 \\
\midrule
\multirow{4}{*}{Assignment} 
& KMeans & \pmstd{0.726}{0.057} & \pmstd{0.656}{0.070} & \pmstd{0.582}{0.078} & \pmstd{0.670}{0.096} & \pmstd{0.772}{0.119} & 0.681 \\
& General OT & \pmstd{0.716}{0.036} & \pmstd{0.652}{0.085} & \pmstd{0.585}{0.082} & \pmstd{0.692}{0.112} & \pmstd{0.769}{0.124} & 0.683 \\
& {w/o curr. $\rho$} & \pmstd{0.720}{0.039} & \pmstd{0.662}{0.083} & \pmstd{0.589}{0.081} & \pmstd{0.686}{0.117} & \pmstd{0.775}{0.126} & 0.686 \\
& \cellcolor{gray!20} \textbf{{w/ curr. $\rho$}} & \cellcolor{gray!20} \pmstd{\textbf{0.726}}{0.039} & \cellcolor{gray!20} \pmstd{\textbf{0.662}}{0.079} & \cellcolor{gray!20} \pmstd{\textbf{0.613}}{0.079} & \cellcolor{gray!20} \pmstd{\textbf{0.685}}{0.131} & \cellcolor{gray!20} \pmstd{\textbf{0.778}}{0.117} & \cellcolor{gray!20} \textbf{0.693} \\
\midrule
\multirow{3}{*}{\cellcolor{white}\makecell{Shared\\ Prototypes}}
& K=16 & \pmstd{0.693}{0.055} & \pmstd{0.659}{0.056} & \pmstd{0.605}{0.065} & \pmstd{0.645}{0.172} & \pmstd{0.747}{0.095} & 0.670 \\
& \cellcolor{gray!20} K=32 & \cellcolor{gray!20} \pmstd{\textbf{0.726}}{0.039} & \cellcolor{gray!20} \pmstd{\textbf{0.662}}{0.079} & \cellcolor{gray!20} \pmstd{\textbf{0.613}}{0.079} & \cellcolor{gray!20} \pmstd{\textbf{0.685}}{0.131} & \cellcolor{gray!20} \pmstd{\textbf{0.778}}{0.117} & \cellcolor{gray!20} \textbf{0.693} \\
& K=50 & \pmstd{0.702}{0.078} & \pmstd{0.658}{0.065} & \pmstd{0.552}{0.062} & \pmstd{0.638}{0.188} & \pmstd{0.763}{0.121} & 0.663 \\
\midrule
\multirow{4}{*}{\cellcolor{white}\makecell{KL-constraint\\ weight}}
& $\gamma{=}1.0$ & \pmstd{0.705}{0.021} & \pmstd{0.659}{0.083} & \pmstd{0.606}{0.078} & \pmstd{0.685}{0.130} & \pmstd{0.774}{0.123} & 0.686 \\
& $\gamma{=}0.3$ & \pmstd{0.731}{0.035} & \pmstd{0.660}{0.084} & \pmstd{0.606}{0.080} & \pmstd{0.684}{0.130} & \pmstd{0.778}{0.117} & 0.692 \\
& $\gamma{=}0.2$ & \pmstd{0.727}{0.037} & \pmstd{0.662}{0.080} & \pmstd{0.606}{0.080} & \pmstd{0.684}{0.131} & \pmstd{0.777}{0.114} & 0.691 \\
& \cellcolor{gray!20} $\gamma{=}0.1$ & \cellcolor{gray!20} \pmstd{\textbf{0.726}}{0.039} & \cellcolor{gray!20} \pmstd{\textbf{0.662}}{0.079} & \cellcolor{gray!20} \pmstd{\textbf{0.613}}{0.079} & \cellcolor{gray!20} \pmstd{\textbf{0.685}}{0.131} & \cellcolor{gray!20} \pmstd{\textbf{0.778}}{0.117} & \cellcolor{gray!20} \textbf{0.693} \\
\midrule
\multirow{4}{*}{\cellcolor{white}\makecell{Instance \\loss weight}}
& $\lambda_{\text{inst}}{=}1.0$ & \pmstd{0.705}{0.040} & \pmstd{0.658}{0.079} & \pmstd{0.605}{0.081} &  \pmstd{0.683}{0.120} & 
\pmstd{0.774}{0.116} & 0.685 \\
& \cellcolor{gray!20} $\lambda_{\text{inst}}{=}0.5$ & \cellcolor{gray!20} \pmstd{\textbf{0.726}}{0.039} & \cellcolor{gray!20} \pmstd{\textbf{0.662}}{0.079} & \cellcolor{gray!20} \pmstd{\textbf{0.613}}{0.079} & \cellcolor{gray!20} \pmstd{\textbf{0.685}}{0.131} & \cellcolor{gray!20} \pmstd{\textbf{0.778}}{0.117} & \cellcolor{gray!20} \textbf{0.693} \\
& $\lambda_{\text{inst}}{=}0.3$ & \pmstd{0.723}{0.051} & \pmstd{0.657}{0.084} & \pmstd{0.603}{0.078} & \pmstd{0.677}{0.122} & 
\pmstd{0.777}{0.112} & 0.687 \\
& $\lambda_{\text{inst}}{=}0.1$ & \pmstd{0.720}{0.039} & \pmstd{0.660}{0.082} & \pmstd{0.586}{0.085} & \pmstd{0.688}{0.111} & 
\pmstd{0.767}{0.129} & 0.684 \\
\bottomrule
\end{tabular}
\end{table*}

\myparagraph{Different OT Types for Instance-to-Prototype Assignment.}
The overall ranking of General OT, UOT without curriculum mass, and UOT with curriculum mass is already summarized in the main paper, and we emphasize the cohort-wise pattern here. Moving from General OT to UOT with curriculum mass brings the largest gain on STAD (\textbf{0.585}$\rightarrow$\textbf{0.613}), while BRCA, BLCA, and KIRC also show modest improvements. CRC behaves differently: General OT is already strong on this cohort, and adding curriculum mass does not further improve the score. This pattern suggests that the curriculum is most useful when early token-to-prototype matches are noisy or diffuse, whereas cohorts with already confident OT assignments benefit less from additional conservatism.

\myparagraph{OT or KMeans for Instance-to-Prototype Assignment.}
Relative to hard KMeans assignment, the soft OT family shows its clearest advantages on STAD and CRC. In particular, the improvement from KMeans to UOT with curriculum mass is \textbf{+3.1\%} on STAD and \textbf{+1.5\%} on CRC. This indicates that hard assignment is especially brittle when tokens are heterogeneous. At the same time, the table shows that General OT already recovers much of the gain on CRC, whereas STAD continues to benefit from curriculum mass. Taken together, these results suggest that the main role of curriculum is not merely to soften assignments, but to delay commitment on cohorts where ambiguous tokens would otherwise be matched too aggressively.

\myparagraph{Multi-Head Mechanism.}
To assess the stability of our multi-head consistency design, we varied the number of heads. We observe that the average performance peaks at $H{=}5$ ($0.693$), whereas $H{=}1$ (single-head) yields $0.669$, $H{=}3$ yields $0.643$ (\textbf{-2.6\%} relative to $H{=}1$), and $H{=}8$ yields $0.681$ (\textbf{-1.2\%} relative to $H{=}5$). These results suggest that a moderate number of heads offers a favorable trade-off among accuracy, stability and efficiency. Fewer heads reduce compute but are more sensitive to view-sampling variance in early training, whereas more heads generally improve stability with diminishing returns and increased training time.

\myparagraph{Number of Shared Prototypes.}
Varying the size of the shared prototype bank shows a clear peak at $K{=}32$ (average $\mathbf{0.693}$). Relative to $K{=}16$ ($0.670$) and $K{=}50$ ($0.663$), the improvements are \textbf{+2.3\%} and \textbf{+3.0\%}. This supports that a medium-sized prototype bank balances expressiveness and transport stability.

\myparagraph{KL-constraint Weight.} Sweeping the KL regularization weight shows that $0.1$ yields the best average $\mathbf{0.693}$. This is slightly higher than $0.3$ ($0.692$, \textbf{+0.1\%}) and $0.2$ ($0.691$, \textbf{+0.2\%}) and clearly higher than $1.0$ ($0.686$, \textbf{+0.7\%}).  A lighter KL pull allows more data-driven prototype usage while keeping sufficient regularization, leading to more stable assignments and better overall performance. 

\myparagraph{Instance Loss Weight.}
Varying the instance-level UOT loss weight $\lambda_{\mathrm{inst}}$ shows that a moderate value provides the best trade-off. The average peaks at $\lambda_{\mathrm{inst}}{=}0.5$ ($\mathbf{0.693}$), improving over $\lambda_{\mathrm{inst}}{=}1.0$ ($0.685$) by \textbf{+1.2\%}, while $0.3$ and $0.1$ yield $0.687$ and $0.684$. These results indicate that too small a weight under-utilizes the token-to-prototype supervision, whereas too large a weight can over-emphasize early pseudo-label noise. A mid-range weight stabilizes optimization and yields the best overall performance.

\myparagraph{Role of the \textsc{Together} Stage.}
Taken together, these ablations clarify the role of the \textsc{Together} stage. Shared prototypes define a common semantic interface across modalities, and the UOT-based soft assignment prevents uncertain tokens from being committed to that interface too early. This interpretation is consistent with the gains of curriculum-based UOT over hard assignment and standard OT: early in training, token locations in the shared space are still unstable, so partial transport and the sink reduce the risk of fixing noisy evidence into the prototype bank. The preference for a moderate prototype number ($K{=}32$) is also consistent with this view. If $K$ is too small, the shared space is overly compressed; if $K$ is too large, it becomes over-fragmented and cross-modal consensus becomes harder to accumulate. Overall, the strongest settings are those that make the prototype-mediated common space both shared and stable.

\begin{figure}[t]
    \centering
    \includegraphics[width=0.9\linewidth]{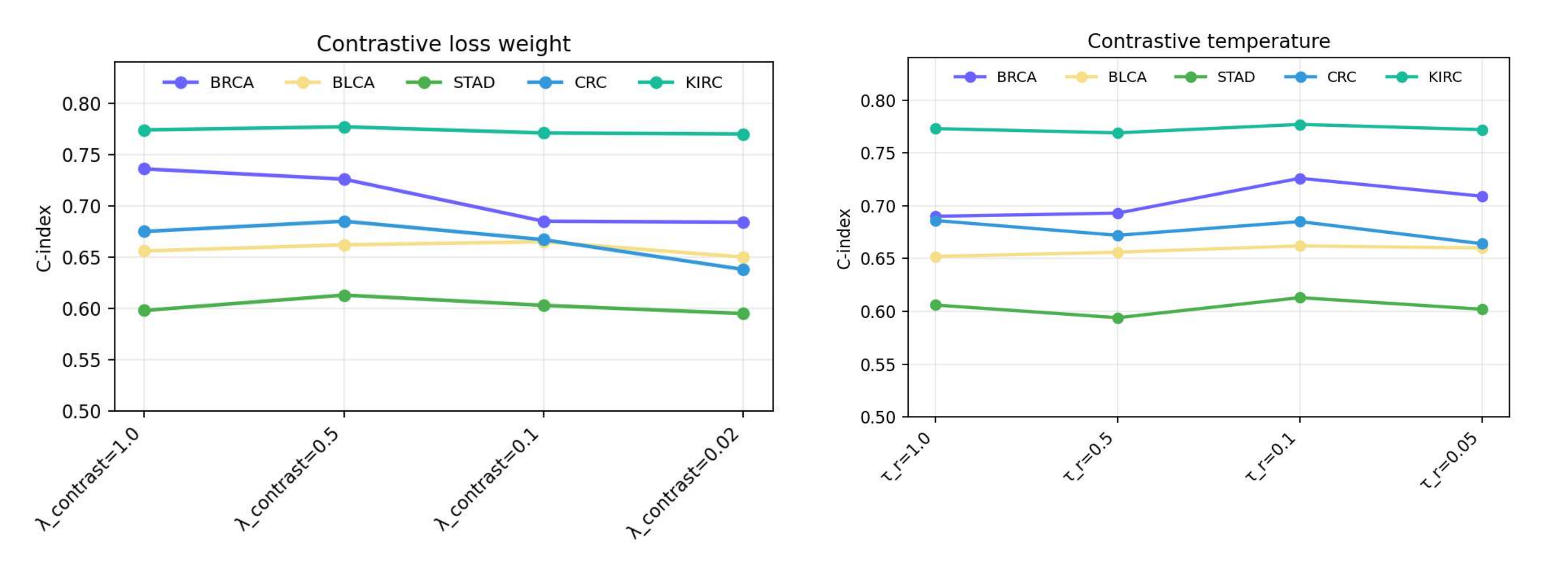}
    \caption{Hyperparameter analysis in the \textsc{Apart} stage.}
    \label{fig:apart ablations}
\end{figure}

\begin{table*}[t]
\setlength{\arrayrulewidth}{0.1mm} 
\scriptsize 
\setlength{\tabcolsep}{1.0pt}
\renewcommand{\arraystretch}{1.00}
\centering
\caption{Hyperparameter experiments in the \textsc{Apart} stage: contrastive temperature and contrastive loss weight. Results are C-index (mean $\pm$ std).}
\label{tab:apart_ablation}
\begin{tabular}{c|c|ccccc|c}
\toprule
\textbf{Module} & \textbf{Settings} & \textbf{BRCA} & \textbf{BLCA} & \textbf{STAD} & \textbf{CRC} & \textbf{KIRC} & \textbf{Avg} \\
\midrule
\multirow{4}{*}{\cellcolor{white}\makecell{Contrastive \\ temperature $\tau_r$}} 
& $1.0$ & \pmstd{0.690}{0.069} & \pmstd{0.652}{0.082} & \pmstd{0.606}{0.067} & \pmstd{0.686}{0.121} & \pmstd{0.773}{0.112} & 0.681 \\
& $0.5$ & \pmstd{0.693}{0.057} & \pmstd{0.656}{0.073} & \pmstd{0.594}{0.090} & \pmstd{0.672}{0.120} & \pmstd{0.769}{0.118} & 0.677 \\
& \cellcolor{gray!20} $0.1$ & \cellcolor{gray!20} \pmstd{\textbf{0.726}}{0.039} & \cellcolor{gray!20} \pmstd{\textbf{0.662}}{0.079} & \cellcolor{gray!20} \pmstd{\textbf{0.613}}{0.079} & \cellcolor{gray!20} \pmstd{\textbf{0.685}}{0.131} & \cellcolor{gray!20} \pmstd{\textbf{0.778}}{0.117} & \cellcolor{gray!20} \textbf{0.693} \\
& $0.05$ & \pmstd{0.709}{0.052} & \pmstd{0.660}{0.084} & \pmstd{0.602}{0.081} & \pmstd{0.664}{0.138} & \pmstd{0.772}{0.125} & 0.681 \\
\midrule
\multirow{4}{*}{\cellcolor{white}\makecell{Contrastive loss \\weight $\lambda_{\text{contrast}}$}}
& $1.0$ & \pmstd{0.736}{0.049} & \pmstd{0.656}{0.076} & \pmstd{0.598}{0.070} & \pmstd{0.675}{0.080} & \pmstd{0.774}{0.120} & 0.688 \\
& \cellcolor{gray!20} $0.5$ & \cellcolor{gray!20} \pmstd{\textbf{0.726}}{0.039} & \cellcolor{gray!20} \pmstd{\textbf{0.662}}{0.079} & \cellcolor{gray!20} \pmstd{\textbf{0.613}}{0.079} & \cellcolor{gray!20} \pmstd{\textbf{0.685}}{0.131} & \cellcolor{gray!20} \pmstd{\textbf{0.778}}{0.117} & \cellcolor{gray!20} \textbf{0.693} \\
& $0.1$ & \pmstd{0.685}{0.051} & \pmstd{0.665}{0.059} & \pmstd{0.603}{0.085} & \pmstd{0.667}{0.128} & \pmstd{0.771}{0.118} & 0.678 \\
& $0.02$ & \pmstd{0.684}{0.065} & \pmstd{0.650}{0.082} & \pmstd{0.595}{0.093} & \pmstd{0.638}{0.151} & \pmstd{0.770}{0.121} & 0.667 \\
\bottomrule
\end{tabular}
\end{table*}

\subsection{Ablations in \textsc{Apart} stage}
We further evaluate the \textsc{Apart} stage to probe the sensitivity of the contrastive refiner. Specifically, we vary the temperature and the contrastive loss weight to assess stability and effectiveness. Table~\ref{tab:apart_ablation} and Fig~\ref{fig:apart ablations} report the results.

\begin{table*}[htb]
\setlength{\arrayrulewidth}{0.1mm} 
\scriptsize 
\setlength{\tabcolsep}{1.0pt}
\renewcommand{\arraystretch}{1.00}
\centering
\caption{Comparisons with SOTA methods of C-index (mean $\pm$ std) when replacing UNI with ResNet50. ``Modal.'' denotes Modality.}
\label{tab:extractor_resnet50}
\begin{tabular}{c|c|ccccc|c}
\toprule
\textbf{Model} & \textbf{Modal.} & \textbf{BRCA} & \textbf{BLCA} & \textbf{STAD} & \textbf{CRC} & \textbf{KIRC} & \textbf{Overall} \\
\midrule
MOTCat~\cite{xu2023multimodal} & g.+ h. & \pmstd{0.671}{0.083} & \pmstd{0.670}{0.036} & \pmstd{0.559}{0.045} & \pmstd{0.622}{0.171} & \pmstd{0.721}{0.134} & 0.649 \\
PIBD~\cite{zhang2024prototypical}   & g.+ h. & \pmstd{0.659}{0.094} & \pmstd{0.653}{0.033} & \pmstd{0.556}{0.072} & \pmstd{0.609}{0.180} & \pmstd{0.725}{0.100} & 0.640 \\
LD-CVAE~\cite{zhou2025robust} & g.+ h. & \pmstd{0.670}{0.072} & \pmstd{0.649}{0.039} & \pmstd{0.590}{0.094} & \pmstd{0.598}{0.136} & \pmstd{0.761}{0.124} & 0.654 \\
MMP~\cite{song2024multimodal}    & g.+ h. & \pmstd{0.706}{0.069} & \pmstd{0.631}{0.049} & \pmstd{0.586}{0.083} & \pmstd{0.613}{0.142} & \pmstd{0.756}{0.122} & 0.658 \\
\rowcolor{gray!20}
\textbf{TTA (Ours)} & \cellcolor{gray!20} g.+ h. & \cellcolor{gray!20} \pmstd{\textbf{0.678}}{0.126} & \cellcolor{gray!20} \pmstd{\textbf{0.662}}{0.043} & \cellcolor{gray!20} \pmstd{\textbf{0.585}}{0.056} & \cellcolor{gray!20} \pmstd{\textbf{0.674}}{0.217} & \cellcolor{gray!20} \pmstd{\textbf{0.787}}{0.097} & \cellcolor{gray!20} \textbf{0.677} \\
\bottomrule
\end{tabular}
\end{table*}

\myparagraph{Weights and Temperature.}
For clarity, we recall that the contrastive temperature $\tau_r$ rescales the logits in the InfoNCE objective in Eq.~(\ref{eq:contrast}), thereby controlling the sharpness of the distinction between positives and negatives. A smaller temperature sharpens the distribution and increases separation, whereas a larger temperature smooths the distribution and emphasizes stability.

\myparagraph{Contrastive Temperature.} On temperature, $\tau_r{=}0.1$ achieves the best average $\mathbf{0.693}$. Relative to $\tau_r{=}1.0$ and $\tau_r{=}0.5$ the gains are \textbf{+1.2\%} and \textbf{+1.6\%}. Very small $\tau_r$ such as $0.05$ drops back to $0.681$. These trends suggest that a moderate temperature yields the most favorable balance between discrimination and robustness, avoiding both over-smoothing and over-sharpening.

\myparagraph{Contrastive Loss Weight.} On the contrastive loss weight, $\lambda_{\text{contrast}}{=}0.5$ delivers the best average $\mathbf{0.693}$. Relative to $\lambda_{\text{contrast}}{=}1.0$ the improvement is \textbf{+0.5\%}, and relative to $0.1$ and $0.02$ the improvements are \textbf{+1.5\%} and \textbf{+2.6\%}. It is worth noting that CRC benefits more from larger contrastive weights, which corroborates the role of our \textsc{Apart} stage in preserving modality‑specific cues. Earlier observations showed that on CRC several multimodal methods lag behind WSI-only MIL baselines, indicating that the survival signal is primarily carried by histopathology and that excessive cross‑modal over-alignment can collapse WSI-specific information. Within our framework, increasing $\lambda_{\mathrm{contrast}}$ in a reasonable range strengthens the modality‑specific regularization, better preserves WSI-specific cues, thus yields higher C‑index on CRC. 

\myparagraph{Role of the \textsc{Apart} Stage.}
Taken together, these results show that the \textsc{Apart} stage is not simply an additional round of contrastive tuning. Its role is to reorganize modality-specific structure after the shared prototype space has been established by \textsc{Together}. Moderate values of $\tau_r$ and $\lambda_{\text{contrast}}$ work best because the anchor signal must be strong enough to recover modality identity, yet not so strong that it breaks compatibility with the shared prototype geometry or amplifies noisy variation. This interpretation is also consistent with the larger sensitivity on CRC, where preserving modality-specific cues is especially important.

\subsection{Other Ablations}
\myparagraph{Image Feature Extractor.}
We also test generalization performance of our method by replacing the underlying image feature extractor UNI~\cite{chen2024uni} with a ResNet50 pretrained on ImageNet~\cite{deng2009imagenet}. Results are shown in Table~\ref{tab:extractor_resnet50}.
With ResNet50 features, TTA still achieves the best Overall, indicating that the benefit of the TTA design is not tied to a single strong pathology encoder. The cohort-wise pattern is also informative: the largest advantages remain on CRC and KIRC, whereas BRCA and STAD become more competitive across methods. This suggests that stronger visual backbones improve the absolute quality of pathology features, but the gain from balancing alignment and distinctiveness persists even when the image encoder is substantially weaker.

\begin{table*}[t]
\setlength{\arrayrulewidth}{0.1mm} 
\scriptsize
\setlength{\tabcolsep}{1.05pt}
\renewcommand{\arraystretch}{1.05}
\centering
\caption{Experiments on key training parameters. Results are C-index (mean $\pm$ std).}
\label{tab:training_params}
\begin{tabular}{c|c|ccccc|c}
\toprule
\textbf{Parameter} & \textbf{Change} & \textbf{BRCA} & \textbf{BLCA} & \textbf{STAD} & \textbf{CRC} & \textbf{KIRC} & \textbf{Avg} \\
\midrule
Bag size & $4096 \rightarrow 2048$ & \pmstd{0.679}{0.036} & \pmstd{0.660}{0.067} & \pmstd{0.604}{0.074} & \pmstd{0.642}{0.126} & \pmstd{0.777}{0.108} & 0.672 \\
Batch size & $32 \rightarrow 16$ & \pmstd{0.682}{0.066} & \pmstd{0.655}{0.059} & \pmstd{0.591}{0.050} & \pmstd{0.662}{0.121} & \pmstd{0.773}{0.103} & 0.673 \\
Batch size & $32 \rightarrow 64$ & \pmstd{0.711}{0.079} & \pmstd{0.649}{0.076} & \pmstd{0.585}{0.056} & \pmstd{0.640}{0.115} & \pmstd{0.770}{0.102} & 0.671 \\
Refiner layers & $1 \rightarrow 2$ & \pmstd{0.712}{0.063} & \pmstd{0.670}{0.065} & \pmstd{0.605}{0.070} & \pmstd{0.688}{0.123} & \pmstd{0.772}{0.096} & 0.690 \\
Learning rate & $1\mathrm{e}{-4} \rightarrow 5\mathrm{e}{-5}$ & \pmstd{0.720}{0.086} & \pmstd{0.658}{0.062} & \pmstd{0.613}{0.075} & \pmstd{0.655}{0.133} & \pmstd{0.779}{0.110} & 0.685 \\
Loss function & Cox $\rightarrow$ NLL & \pmstd{0.667}{0.176} & \pmstd{0.642}{0.093} & \pmstd{0.576}{0.076} & \pmstd{0.698}{0.141} & \pmstd{0.803}{0.077} & 0.677 \\
Max epochs & $30 \rightarrow 50$ & \pmstd{0.728}{0.059} & \pmstd{0.661}{0.076} & \pmstd{0.605}{0.083} & \pmstd{0.678}{0.137} & \pmstd{0.770}{0.127} & 0.688 \\
$\rho$ ramp-up & Sigmoid $\rightarrow$ Linear & \pmstd{0.735}{0.072} & \pmstd{0.661}{0.060} & \pmstd{0.604}{0.053} & \pmstd{0.667}{0.116} & \pmstd{0.771}{0.113} & 0.687 \\
\bottomrule
\end{tabular}
\end{table*}
\myparagraph{Training Parameters.}
Table~\ref{tab:training_params} complements the robustness summary in the main paper by unpacking the effect of several training choices. Reducing the WSI bag size or the batch size causes moderate degradation, which is consistent with noisier slide summaries and, for Cox loss, weaker risk-set estimation. Increasing the refiner depth from one to two layers remains competitive ($0.690$), suggesting that the \textsc{Apart} module is not highly sensitive to small architectural changes. Replacing Cox with discrete-time NLL or extending training to 50 epochs also preserves most of the performance. As summarized in the main paper, the sigmoid schedule remains preferable to a linear $\rho$ ramp-up; Table~\ref{tab:training_params} shows that this difference is modest rather than catastrophic. Overall, the results vary smoothly around the default configuration, suggesting that the best setting is not an isolated optimum produced by a fragile hyperparameter choice.

\myparagraph{Robustness Across Encoders, Objectives, and Training Settings.}
Taken together, the results in this subsection support a robustness interpretation beyond standard sensitivity analysis. Replacing UNI with ResNet50 shows that the gain is not driven solely by a specific image encoder. Varying the training recipe indicates that the method does not depend on a single narrow configuration, and replacing Cox with discrete-time NLL shows that the improvement is not tied to one favorable survival objective. Although the default setting remains best overall, the surrounding variants stay competitive, suggesting that TTA is not brittle and that its gains arise from the model design rather than a single confounding recipe.

\section{More Visualizations and Analysis}
\label{sec:tta_additional_visualization}

We next present additional visual analyses of the learned prototype space, slide-level assignment patterns, and case-specific prototype-pathway associations. Together, these figures help clarify how the shared prototypes organize histologic patterns and how they relate to pathway-level genomic signals.

\begin{figure}[t]
    \centering
    \includegraphics[width=0.6\linewidth]{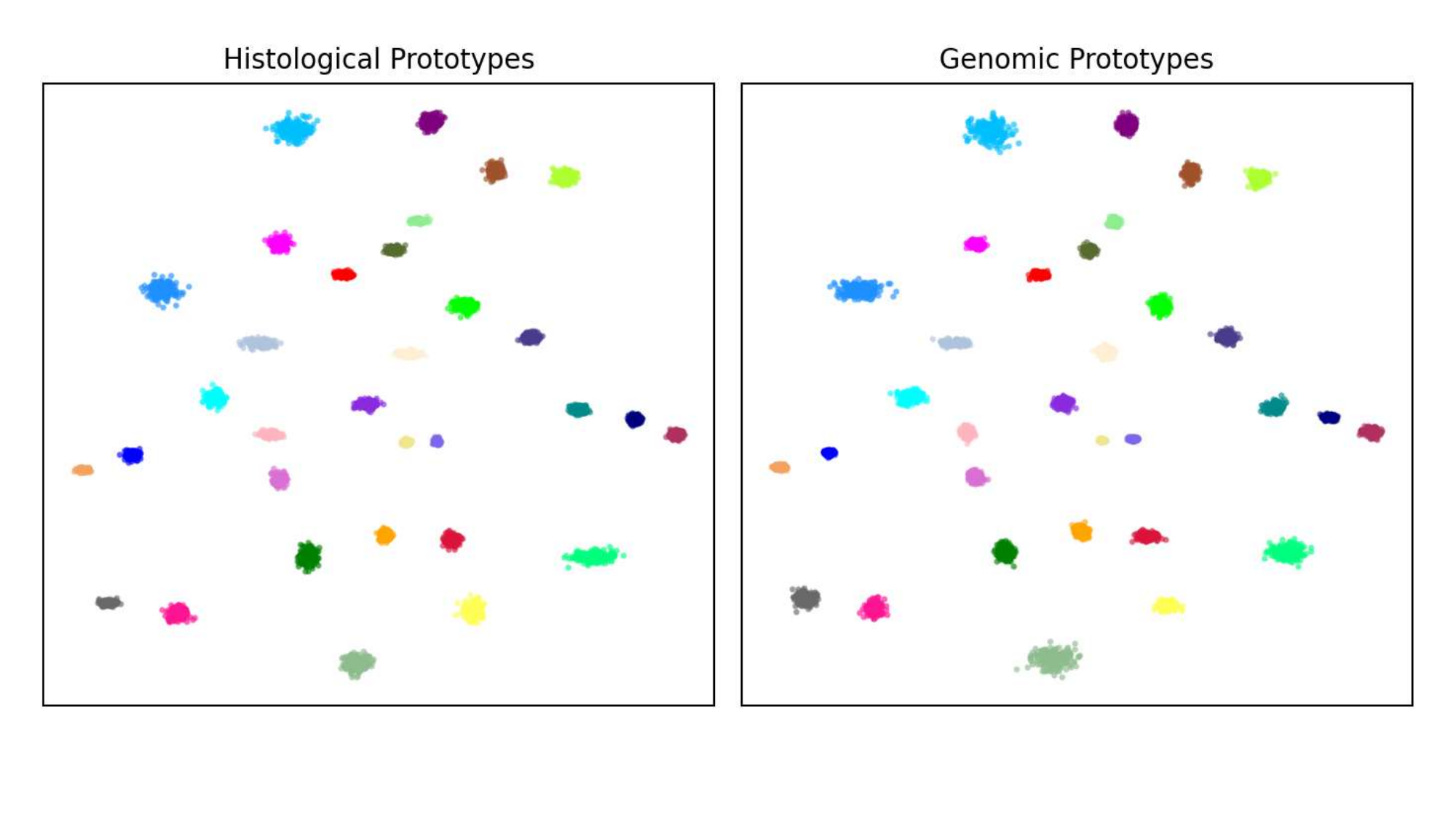}
    \caption{Visualization of the learned shared prototypes.}
    \label{fig:prototypes_visualization}
\end{figure}

\begin{figure*}[t]
    \centering
    \includegraphics[width=0.95\linewidth]{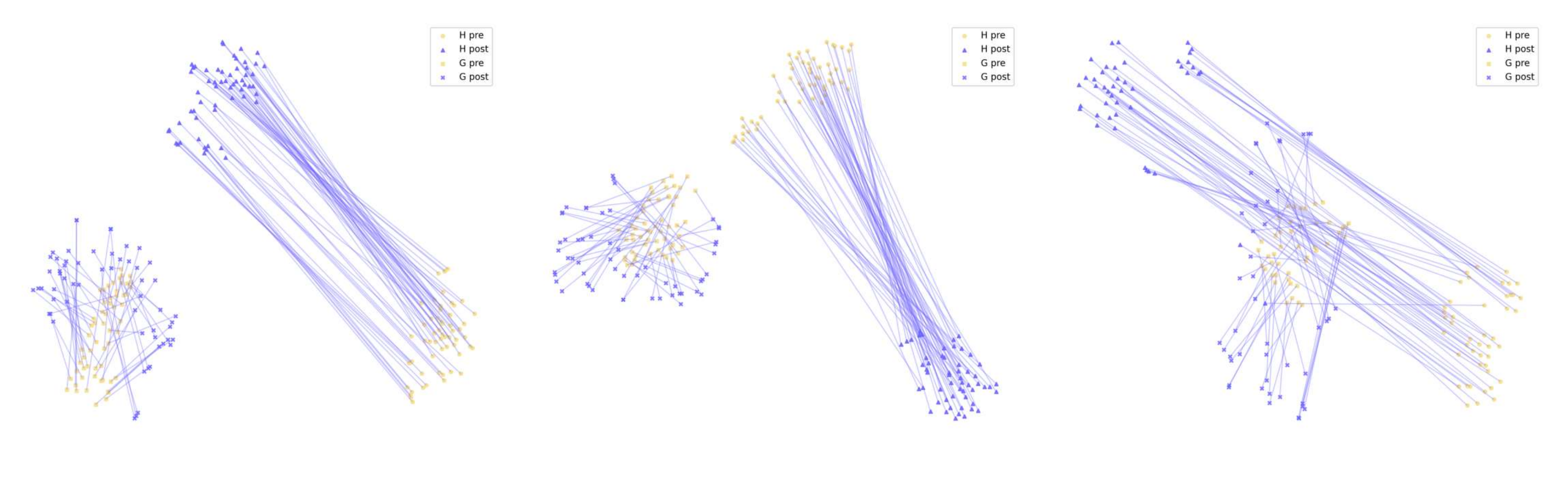}
    \caption{t‑SNE of modality-specific token embeddings before and after the \textsc{Apart} stage.}
    \label{fig:bef-aft-apart}
\end{figure*}

\begin{figure}[t]
    \centering
    \includegraphics[width=0.98\linewidth]{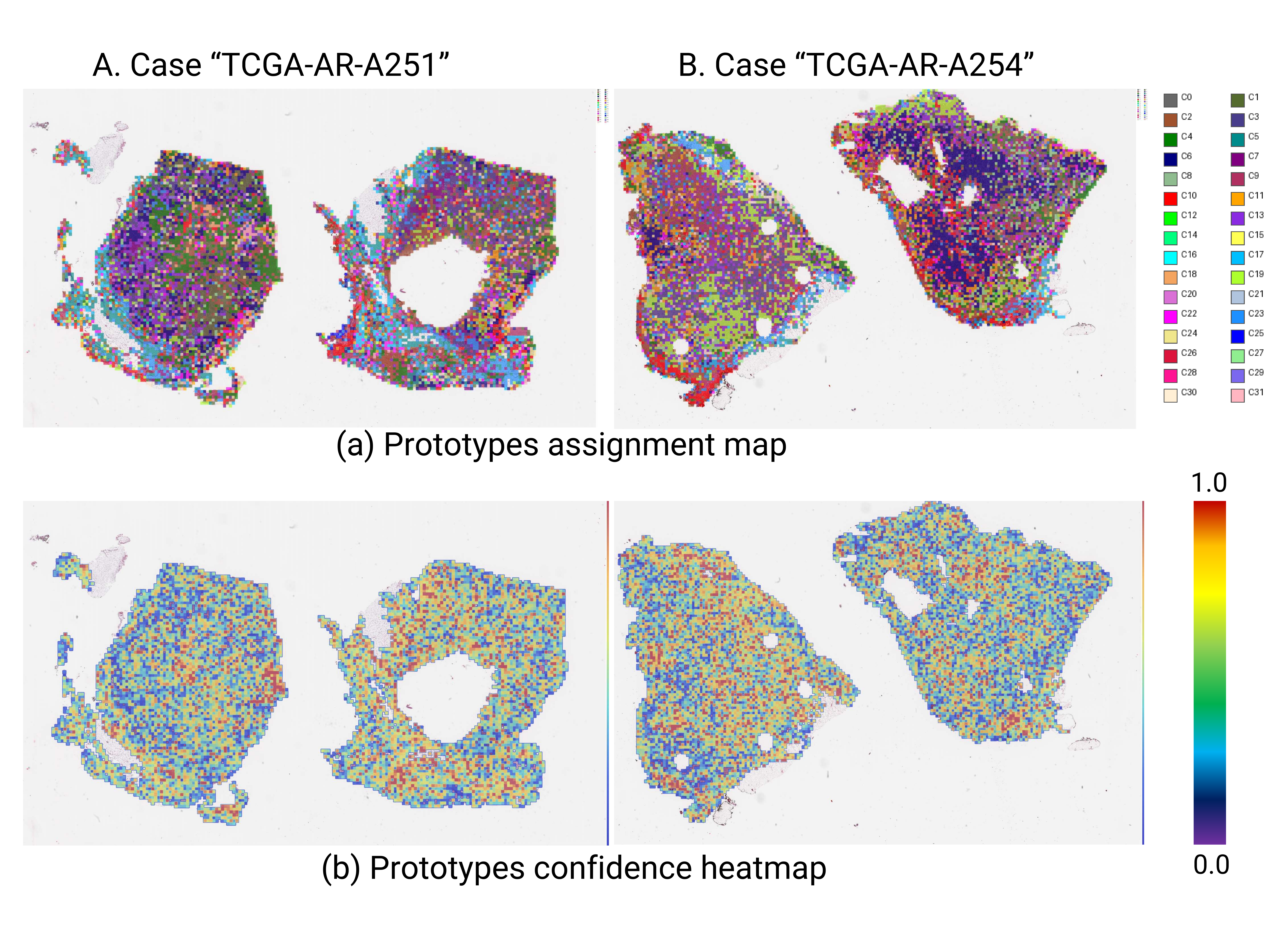}
    \caption{\textbf{Prototype assignment and confidence maps for two BRCA cases.} Columns \textbf{A} and \textbf{B} correspond to Cases \texttt{TCGA-AR-A251} and \texttt{TCGA-AR-A254}, respectively. \textbf{(a)} shows the prototype-assignment map, where each patch is colored by the prototype with the highest posterior probability. \textbf{(b)} shows the corresponding prototype-confidence heatmap, computed as the maximum posterior probability over all prototypes at each patch.}
    \label{fig:wsi_heatmaps}
\end{figure}

\myparagraph{Shared Prototypes Visualization.}
The shared prototypes are visualized in Fig~\ref{fig:prototypes_visualization}. We project the learned prototype bank into a 2-dimensional space with t-SNE and display one color per prototype. The prototypes form separated yet related groups in the projected space, suggesting that the bank does not collapse into a few redundant anchors. This structured organization is consistent with the role of the \textsc{Together} stage: it learns a prototype-mediated common space that is shared across modalities while preserving prototype-level specialization.

\begin{figure*}
    \centering
    \includegraphics[width=1\linewidth]{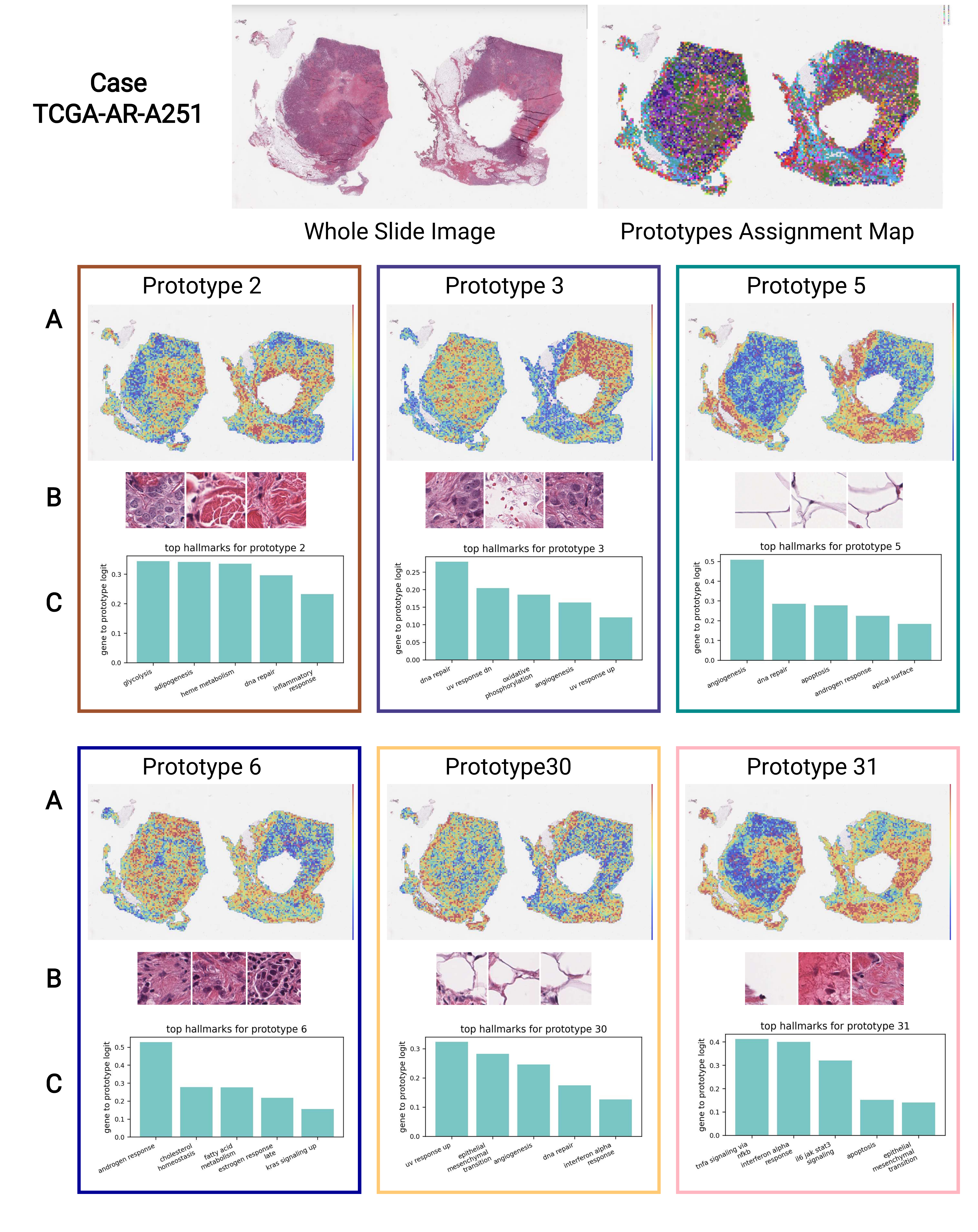}
    \caption{\textbf{Prototype-centered interpretability visualization for Case \texttt{TCGA-AR-A251}.} The top row shows the whole-slide image and the prototype-assignment map. Six case-specific prototypes are highlighted. For each prototype block, row \textbf{A} shows the prototype-specific WSI heatmap, row \textbf{B} shows top-3 representative high-scoring patches, and row \textbf{C} shows the top hallmark pathways ranked by gene-to-prototype logits. The colored border of each prototype block matches the corresponding prototype color in the assignment map.}
    \label{fig:case_251}
\end{figure*}

\begin{figure*}
    \centering
    \includegraphics[width=1\linewidth]{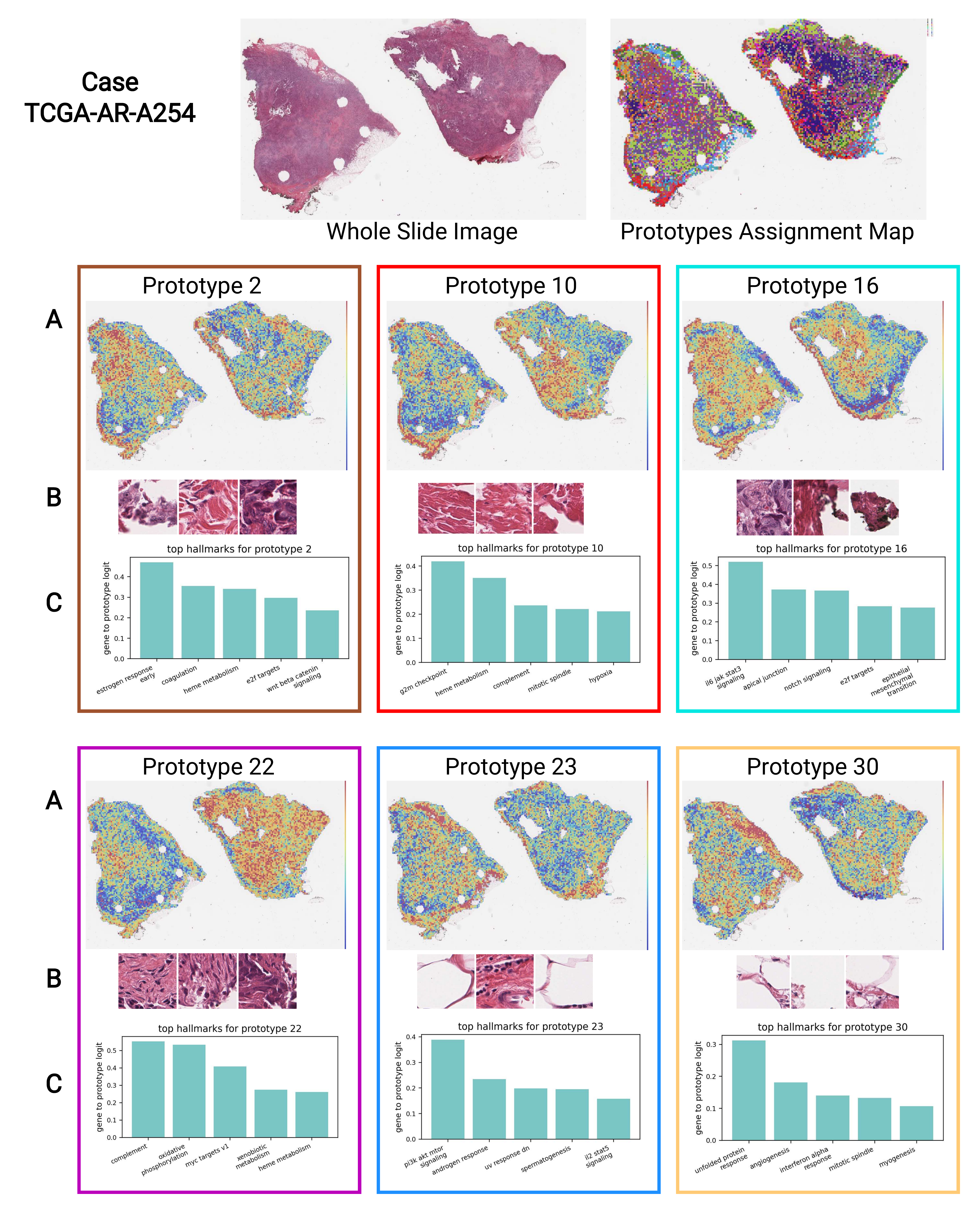}
    \caption{\textbf{Prototype-centered interpretability visualization for Case \texttt{TCGA-AR-A254}.} The top row shows the whole-slide image and the prototype-assignment map. Six case-specific prototypes are highlighted. For each prototype block, row \textbf{A} shows the prototype-specific WSI heatmap, row \textbf{B} shows top-3 representative high-scoring patches, and row \textbf{C} shows the top hallmark pathways ranked by gene-to-prototype logits. The colored border of each prototype block matches the corresponding prototype color in the assignment map.}
    \label{fig:case_254}
\end{figure*}

% \begin{figure*}
%     \centering
%     \includegraphics[width=0.98\linewidth]{fig/prototypes-heatmaps.pdf}
%     \vspace{-3mm}
%     \caption{Heatmaps of 32 prototypes.}
%     \label{fig:prototypes_heatmaps}
%     \vspace{-2mm}
% \end{figure*}

\myparagraph{Before and After the \textsc{Apart} Stage.}
Fig~\ref{fig:bef-aft-apart} shows t-SNE of modality-specific token embeddings before and after the \textsc{Apart} stage. After refinement, histopathology tokens and genomic tokens show consistent motion toward their modality-specific clusters, suggesting that they systematically shift toward their respective anchors. This figure therefore reflects a more ordered modality-specific regrouping, rather than merely a larger inter-modal gap.

\myparagraph{Prototype-centered Interpretability Visualization.}
Fig.~\ref{fig:wsi_heatmaps} provides a slide-level summary of two representative BRCA cases. The prototype-assignment maps visualize the dominant prototype at each patch location, whereas the confidence heatmaps quantify how decisively each location is explained by a single prototype. Low-confidence regions often correspond to transition zones or morphologically mixed tissue, giving a natural boundary to the interpretation rather than forcing certainty everywhere.

Figs.~\ref{fig:case_251} and~\ref{fig:case_254} show that the selected prototypes are not activated diffusely across the WSI. Instead, each prototype occupies a relatively localized spatial territory, and different prototypes highlight different regions within the same case. This interpretation is further supported by Fig.~\ref{fig:wsi_heatmaps}: the assignment map reveals a global prototype partition over the slide, while the confidence map shows that these assignments are more decisive in some regions than in others, which is consistent with the presence of morphologically coherent areas as well as transitional or mixed regions. Taken together, these observations suggest that the learned shared prototypes form a structured slide-level partition rather than a collection of arbitrary latent centroids.

The top-3 patches are included to make this spatial organization histologically interpretable. Concretely, they are the highest-scoring patches associated with a given prototype, so they reveal what local morphology gives rise to the activation seen in the prototype-specific heatmap. This visual evidence is important because the heatmap alone indicates where a prototype is active, but not what tissue pattern it corresponds to. Showing multiple high-scoring examples, rather than a single patch, also makes it possible to assess whether a prototype is visually coherent across different spatial locations. In both cases, patches associated with the same prototype exhibit recurrent visual characteristics, suggesting that the prototype captures a coherent histomorphologic pattern rather than isolated patch-level artifacts.

Importantly, each prototype is also paired with its top associated hallmark pathways in the genomic modality. Although these pathway profiles should be interpreted as associations rather than causal mechanisms, they show that the same prototype that organizes a spatial tissue pattern is also linked to a non-random molecular program. Taken together, rows \textbf{A}, \textbf{B}, and \textbf{C} suggest that a prototype functions as a multimodal semantic anchor: it is spatially localized on the slide, visually instantiated by recurrent histologic patches, and linked to a structured set of pathway-level genomic signals.

% \begin{figure}
%     \centering
%     \includegraphics[width=0.98\linewidth]{fig/gene_proto.pdf}
%     \vspace{-3mm}
%     \caption{Gene-prototype interactions for one sample. (a) Top‑8 prototypes per pathway token. Each square is split into eight vertical strips that indicate the eight shared prototypes with the highest gene-to-prototype weights for that token. (b) Pathway‑token counts per shared prototype.}
%     \label{fig:gene_prototypes}
%     \vspace{-2mm}
% \end{figure}

\end{document}